\documentclass{article} 
\usepackage{iclr2027_conference,times}

\usepackage{amsmath,amsfonts,bm}

\def\eqref#1{equation~\ref{#1}}

\def\1{\bm{1}}

\DeclareMathAlphabet{\mathsfit}{\encodingdefault}{\sfdefault}{m}{sl}
\SetMathAlphabet{\mathsfit}{bold}{\encodingdefault}{\sfdefault}{bx}{n}

\newcommand{\R}{\mathbb{R}}

\DeclareMathOperator*{\argmin}{arg\,min}

\iclrfinalcopy

\usepackage{setspace}
\usepackage{hyperref}
\usepackage{url}
\usepackage{longtable}

\usepackage{xcolor}

\usepackage{booktabs} 
\usepackage{enumerate} 
\usepackage{comment}
\usepackage{graphicx}
\usepackage{tikz}
\usepackage{siunitx}
\usetikzlibrary{arrows}
\usetikzlibrary{positioning}
\usetikzlibrary{shapes}
\usetikzlibrary{fit}
\usetikzlibrary{backgrounds}
\usepackage{amssymb}
\usepackage[most]{tcolorbox}
\usepackage{tabularx}
\tcbuselibrary{breakable, skins}
\usepackage{listings}
\newtcolorbox{responsebox}[1]{
  title=#1,
  fonttitle=\bfseries\small,
  colback=white,
  colframe=gray!50,
  boxrule=0.5pt,
  breakable,
  before upper={\small}
}

\usepackage[utf8]{inputenc} 
\usepackage[T1]{fontenc}    
\usepackage{hyperref}       
\usepackage{url}            
\usepackage{booktabs}       
\usepackage{amsfonts}       
\usepackage{nicefrac}       
\usepackage{microtype}      
\usepackage{amsmath}

\usepackage{graphicx}
\usepackage{mathtools}
\usepackage{algorithm}
\usepackage{algpseudocode}
\usepackage{subcaption}
\usepackage{caption}
\usepackage{wrapfig}

\usepackage{array}
\usepackage{ragged2e}
\usepackage{multirow}
\usepackage{multicol}
\usepackage{enumitem}
\usepackage{makecell}
\usepackage{verbatim}
\usepackage{parskip} 
\usepackage{soul}
\usepackage{dashrule}
\usepackage{seqsplit}

\definecolor{g}{HTML}{228833}
\definecolor{r}{HTML}{cc6677}
\definecolor{b}{HTML}{2b67c3}
\renewcommand*{\epsilon}{\varepsilon}%

\newcommand{\Eff}{\mathcal{E}}

\title{Distributed Sparse Interventions in Language Models}

\author{%
Maximilian S. Ernst$^{1,2,3}$\thanks{Correspondence to \texttt{ernst@mpib-berlin.mpg.de}. Code is available on \href{https://github.com/Explainable-AI-Berlin/DSI}{GitHub}.
} 
\quad Lorenz Linhardt$^{3,4}$ \quad Aaron Peikert$^{2}$ \quad Oliver Eberle$^{3,4}$ \\
$^1$Max Planck School of Cognition, Leipzig, Germany \\
$^2$Center for Lifespan Psychology, Max Planck Institute for Human Development, Berlin, Germany \\
$^3$Machine Learning Group, Technische Universität Berlin, Berlin, Germany\\
$^4$Berlin Institute for the Foundations of Learning and Data (BIFOLD), Berlin, Germany\\
}

\begin{document}

\maketitle

\begin{abstract}
Language models can perform a wide range of tasks at various levels of abstraction. 
They flexibly infer tasks from context, execute multiple tasks simultaneously, and select among competing goals.
Understanding and controlling how models represent these tasks is central to aligning them with human objectives.
Researchers can change model behaviour in many ways, but model steering, which manipulates intermediate task representations, links representation to behaviour most directly.
Current steering methods typically intervene densely and assume linear effects, e.g., shifting every activation along a global direction in the residual stream.
Because such interventions change thousands of neurons at once, they leave researchers to separate relevant from irrelevant neurons.
We introduce Distributed Sparse Interventions (DSI), which steers models through a sparse set of individual neurons distributed across layers and attention heads.
Finding such a set requires accounting for the selected neurons' interactions, even across layers; DSI achieves that by iteratively refining which neurons to intervene on and how strongly.
Across 12 tasks and three instruction-tuned models, DSI activates task behaviour by identifying and intervening on fewer than 0.1\% of neurons, outperforming dense steering vectors and often matching 10-shot in-context learning.
Because DSI identifies explicit sets of neurons, it connects neuron-level computation to changes in latent representations and, ultimately, to desired outputs, allowing us to study how distributed computational components interact to produce steered behaviour.
Our results demonstrate that steering can be viewed not only as a linear shift in representation space, but also as a targeted intervention on model computation, realised by adjusting a small set of interacting neurons whose coordinated activity spans multiple layers and contributes to behaviour.

\end{abstract}

\section{Introduction}
\begin{figure}[ht]
{\centering
    \includegraphics[width=\textwidth]{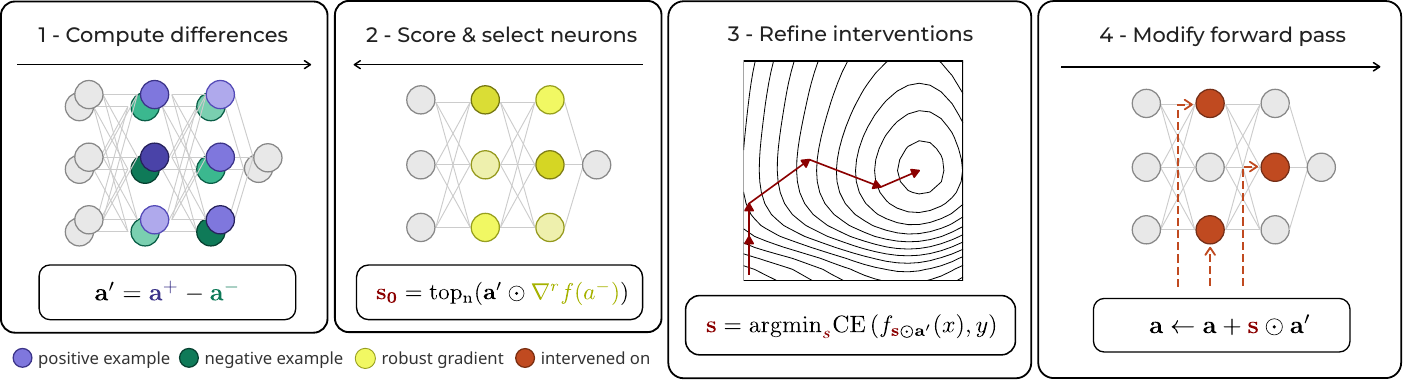}
}
\vspace{-10pt}
\caption{Schematic overview of how distributed sparse interventions are identified and applied via DSI. (1) Activation differences are extracted from positive and negative examples. (2) An initial sparse intervention is computed using a robustified gradient estimate. (3) The intervention is refined via an iterative procedure. (4) The refined intervention is applied additively to modify model behaviour.} \label{fig:overview}
\end{figure}

Modern language models exhibit flexible task-solving capabilities; through in-context learning (ICL), they infer tasks from context, execute multiple tasks simultaneously, and select among competing goals~\citep{icl, li2024language, multitask}.
Understanding and controlling how models represent these tasks is essential to safeguard models and ensure alignment with human objectives. 
Researchers can change model behaviour in several ways, most commonly through prompt engineering and fine-tuning.
Model steering offers a third approach by intervening on intermediate representations that modify the model's forward pass to activate or suppress specific behaviours.
Because steering manipulates task representations, it links representation to behaviour most directly, making it well suited for both flexible, context-dependent adaptation and studying how tasks are represented.
Current steering methods typically rest on three assumptions.
First, they assume that intervention effects are approximately linear: most methods identify directions in activation space~\citep{CAA, FV, taskvectors, beaglehole2026toward, ICV, wu2025axbench}, leveraging the observation that many task-relevant behaviours exhibit approximately linear structure.
Second, they assume that interventions are dense: whether a vector is estimated directly at a specific layer \citep{CAA, taskvectors} or composed from sparse dictionary features \citep{templeton2024scaling}, the resulting intervention modifies large numbers of neurons simultaneously, since features that are sparse in a learned dictionary are dense in the neuron basis~\citep{bricken2023towards, elhage2022toy}.
Recent work has begun to restrict interventions to selected attention heads~\citep{lofit} or subsets of neurons~\citep{feng2026finegrained}, showing that sparser interventions can be effective.
Third, interactions between components are often left out: many methods intervene at a single, predefined layer \vadjust{\nobreak} \citep{CAA, FV}, and methods that optimise interventions typically capture interactions only within that layer or include whole attention heads~\citep{wu2025axbench, reps, lofit}.
These assumptions limit analysis to whole directions or components, leaving researchers to separate relevant from irrelevant neurons.
We introduce \emph{Distributed Sparse Interventions} (DSI), which relaxes all three assumptions.
By iteratively refining which neurons to intervene on and how strongly, DSI captures nonlinear effects and interactions between neurons, while restricting the intervention to a sparse set distributed across layers and attention heads (Figure~\ref{fig:overview}). Our contributions are as follows:

\begin{enumerate}[topsep=1pt, partopsep=0pt, itemsep=0pt, leftmargin=0.7cm]
\renewcommand{\labelenumi}{(\theenumi)}
\item An analysis of neuron-level interventions leading to DSI, a method enabling task activation by identifying and intervening on sparse sets of neurons (Section \ref{sec:dsi}).
\item A demonstration of complex, non-linear dependencies among neuron interventions, highlighting the importance of iterative approaches to determine multi-layer interventions (Section \ref{sec:experiments}).
\item Across 12 tasks~\citep{FV} and three models (Qwen3 (8B)~\citep{qwen3}, Gemma 3 (4B)~\citep{gemma3}, Llama 3.2 (3B)~\citep{llama3}), intervening on 1--64 neurons (fewer than 0.1\% of attention-head neurons) typically suffices to activate a task, outperforming dense function vectors~\citep{FV} and often matching 10-shot ICL (Section~\ref{sec:eval}, Table~\ref{table:main}; Section \ref{sec:sparse}), Figure \ref{fig:acc}.
\item At small budgets, interventions distributed across layers clearly outperform interventions restricted to a single layer (Section~\ref{sec:distributed}, Figure~\ref{fig:sl}).
\item On AxBench~\citep{wu2025axbench}, a benchmark for steering several hundred concepts in open-ended text generation, DSI performs on par with recent steering methods at larger budgets, and clustering its interventions shows that similar neuron sets steer semantically related concepts (Section~\ref{sec:axbench}, Figure \ref{fig:clustering}).
\end{enumerate}
Together, these results challenge the assumption that steering requires dense, linear interventions at specific layers, and show the potential of sparse neuron sets as a more interpretable unit for the study of task representations. Our code will be made  openly available upon publication.

\section{Related work}

\paragraph{Model steering}
Prior work on model steering has largely focused on identifying linear directions in activation space to intervene on model behaviour. Associated vectors are typically calculated from in-context-learning examples~\citep{FV, taskvectors, ICV} as well as contrastive sample pairs~\citep{CAA}. More recently, methods have been proposed that identify vectors via optimisation, such as ReFT-r1 \citep{wu2025axbench}, combining concept detection and steering, or RePS \citep{reps}, proposing a novel preference-optimisation objective. Moving towards sparse interventions, LoFiT~\citep{lofit} trains additive biases applied to specific attention heads, while \citet{feng2026finegrained} show that restricting interventions to a subset of neurons can increase steering performance. Moreover, conditional interventions have been explored in works such as HyperSteer \citep{hypersteer}, a hypernetwork-based approach computing steering vectors dependent on the steered model's representations. In contrast, we propose to intervene on a sparse set of neurons, modelling their nonlinearities and interactions across attention heads and layers.

\paragraph{Sparse representations in neural networks}
Often inspired by neuroscientific findings of sparse coding~\citep{hippocampus, sparsecoding, visualcoding}, prior work on understanding representations in artificial neural networks has investigated sparse representations.
Research on model compression and pruning \citep{li2024discovering}, as well as the lottery ticket hypothesis \citep{frankle2018the, lottery2020}, suggests the existence of underlying sparse structure in large deep learning models by identifying sparse subnetworks capable of matching the performance of dense models.
In vision models, neuron units aligning with human-interpretable concepts were identified \citep{bau2017dissect}. 
Similarly, in language models, specific `skills'~\citep{skillneurons} and `knowledge'~\citep{dai2022knowledge} have been localised to neurons, and sparse subnetworks have been shown to support strong performance through targeted fine-tuning~\citep{grafting}. 
As neural models may represent features in superposition~\citep{elhage2022toy}, recent interpretability work has explored dictionary learning to disentangle representations into interpretable features~\citep{bricken2023towards}, assuming the native neuron basis is not sufficiently sparse. Further efforts have focused on learning a sparse basis of interpretable features either at a specific layer (e.g., sparse autoencoders~\citep{bricken2023towards, huben2023sparse}) or across multiple layers (e.g., crosscoders~\citep{lindsey2024crosscoders}) in order to generate sparser circuits~\citep{marks2025sparse}. While this line of work relies on dense features in high-dimensional spaces, it remains an open question to what extent task-relevant structure is accessible within the native neuron basis. More recently, it has been found that for some circuits previously analysed with sparse autoencoders, circuits of similar sparsity can be recovered in the neuron basis of the MLP components of transformer blocks~\citep{sparsecircuits}.  

\section{DSI - Distributed Sparse Interventions} \label{sec:dsi}

We introduce \emph{Distributed Sparse Interventions} (DSI), a method to identify and apply effective interventions on small sets of neurons distributed across layers and attention heads. As illustrated in Figure~\ref{fig:overview}, DSI follows a four-step procedure: (1) \textbf{activation differences} between positive and negative examples are computed for all neurons to isolate task-specific model responses; (2) a sparse \textbf{initial intervention} is derived from these activation differences and a robustified gradient; (3) a \textbf{refined intervention} is generated via step-wise optimisation of the sparse neuron set and neuron-level intervention scaling, accounting for neuron interactions (Appendix ~\ref{app:toy} provides an illustration and analysis demonstrating the necessity of modeling these interactions); and (4) during a \textbf{modified forward pass}, this intervention is applied to elicit the task-solving capabilities on 0-shot examples.

After discussing preliminaries, first-order intervention effect estimation and robustified gradients are defined in Section~\ref{sec:taylor}, and the iterative intervention refinement procedure is introduced in Section~\ref{sec:iterative}.

\paragraph{Notation}
Let $f$ denote a model mapping a sequence $x$ of input tokens to output logits. 
A task is defined via a set $D$ of input-output pairs $(x, y)$. From $D$, we construct positive examples $D^+ = \{(x^+, y^+)\}$ where the model shows the desired behaviour, e.g., via prompting it to solve a certain task or providing in-context examples. Similarly, we construct negative examples $D^- = \{(x^-, y^-)\}$ where a model does not show the desired behaviour, e.g., 0-shot examples or prompts unrelated to the given task. For an input $x$, $f(x)_{y^+}$ refers to the output logit of the correct token, i.e., the first token of $y^+$.

We let $a \in \R$ denote a neuron activation in a hidden layer of $f$ and $\delta \in \R$ an intervention on that neuron. $f_\delta \coloneq f(x|a\gets a+\delta)$ then denotes the model modified by intervention $\delta$, that is, during a forward pass, $a$ is computed from the input, then $\delta$ is added to $a$, and the forward pass is continued. $\mathbf{a} \in \R^d$ denotes the vector of all hidden activations that are considered for an intervention, and $\Delta \in \R^d$ the vector of all interventions.

\subsection{Approximating intervention effects}\label{sec:taylor}

\paragraph{Taylor approximation} To identify which neurons to intervene on for task $D$, we consider the effect of intervening on a neuron during a forward pass in the negative setting. That is, for a prompt $x^-$, the effect of adding an intervention $\delta$ to a neuron $a$ during the forward pass is
\begin{equation}
\Eff(\delta) = f(x^- \vert a \gets a + \delta)_{y^+} - f(x^-)_{y^+},
\end{equation}
i.e., the change in the correct output logit caused by the addition of $\delta$ to $a$ during the forward pass. Since an exact calculation would require an independent forward pass for each neuron in $f$, we approximate the intervention effect via a first-order Taylor approximation:
\begin{align*}
f(x^- \vert a \gets a + \delta)_{y^+} &\approx f(x^-)_{y^+} + \partial_a f(x^-)_{y^+} \; \delta \\
\implies \Eff(\delta) &\approx \partial_a f(x^-)_{y^+} \; \delta
\end{align*}
This is similar to the setting of attribution patching \citep{atp}, with the difference that the activation of $a$ is not replaced during the forward pass (a "structural" intervention) but a term $\delta$ is added to it (a "parametric" intervention) \citep{interventions}. 

\paragraph{Robustified gradient} \label{sec:lrp}
Deep neural network gradients suffer from instabilities such as gradient shattering \citep{shatteredgradients}, making them a potentially unreliable signal for the expected changes in model output under interventions. To mitigate this brittle behaviour, various other methods, such as integrated gradients \citep{integratedgradients}, smooth grad \citep{smoothgrad}, or layer-wise relevance propagation (LRP) \citep{lrp, lrpbook} have been proposed for computing more robust input attributions. 
Recent work extends these findings to activation patching and shows that intermediate attributions, using LRP-modified gradients, better estimate patching effects in transformer-based language models than raw gradients \citep{relp}.
To improve the approximation of the intervention effect, we therefore use an LRP-modified gradient. The modifications include using the identity rule \citep{NEURIPS2024_d6d0e41e} for nonlinear activation functions, the half rule for gated MLPs and attention modules \citep{lrplstm, attnlrp}, and the LayerNorm rule for layer normalization modules \citep{transformerlrp}. We denote this robustified gradient by $\partial^r$ and provide additional details on the rules in Appendix~\ref{app:lrp}.

\paragraph{Average effect} Equipped with this robustified gradient, we approximate the average effect $\bar \Eff$ of the intervention $\delta$ across the dataset of negative examples as 
\begin{equation}
    \bar \Eff(\delta) \approx \frac{1}{\lvert D^- \rvert} \sum_{x^- \in D^-} \partial^r_a f(x^-)_{y^+} \; \delta.\label{eq:app}
\end{equation}

\subsection{Determining the interventions} \label{sec:iterative}

\paragraph{Initial set} \label{sec:initial}
To determine the magnitude of interventions, we follow prior approaches that construct steering signals via activation differences \citep{CAA, FV} and compute the average activation difference of neuron $a$ between a positive and a negative setting during the forward pass:
\begin{equation}
a' \coloneq \frac{1}{\lvert D^+ \rvert} \sum_{x^+ \in D^+} (a\vert x^+) - \frac{1}{\lvert D^- \rvert} \sum_{x^- \in D^-} (a\vert x^-).
\end{equation}

The resulting quantity captures how a neuron’s activation shifts between settings where the task is solved and settings where it is not, providing an informative initialization for intervention strengths.
For a given number of neurons $n$ to intervene on, we then choose the neurons with the largest expected effect $\bar \Eff(a')$ according to Equation \ref{eq:app}. Introducing a scaling factor $\mathbf s \in \R^d$ which is one for the $n$ largest effects and zero otherwise, we write the vector of interventions as $\Delta = \mathbf{s} \odot \mathbf{a'}$.

\paragraph{Iterative refinement}
So far, we approximated the expected effect of an intervention $\delta$ on a neuron $a$ with a first-order Taylor expansion around the root point $(a\vert x^-)$, i.e., the value the neuron takes on during a forward pass on a 0-shot prompt. This disregards nonlinear effects of interventions, as well as interaction effects among interventions, which, as we show in Section~\ref{sec:experiments}, can be sizeable in highly nonlinear transformer networks. As explicit computation of higher-order terms across thousands of interventions is computationally infeasible, we opt for an iterative approximation. 
When maximizing a nonlinear function, 
it is typically not approximated only once at the initial point, but the function is repeatedly approximated, followed by a step in the gradient direction, and this process is iterated until a stopping criterion is met.
Appendix \ref{app:toy} contains an illustrative example showing how such an iterative approach can help model interactions between neuron-level interventions.

By first constructing a candidate set of $n$ neurons with the largest expected effect $\bar \Eff(a')$ (as described above), and iteratively refining the intervention, we arrive at our proposed Algorithm \ref{alg:rel}: for a given neuron set size $n$ and datasets $D^-$ and $D^+$, it (1) computes the average activation differences, (2) computes a robustified gradient w.r.t. the activations in the 0-shot setting, (3) selects the $n$ interventions with the highest expected effect as a starting point and, (4) iteratively refines the effects $t$-times by re-evaluating the expected effects at the current intervention.

For the iterative optimisation steps, we employ ZeroFPR \citep{zerofpr}, an algorithm suited for optimising nonconvex functions with a limited number of nonzero parameters (the per-neuron interventions). Instead of maximising the raw output logit corresponding to the correct answer, we minimise the cross-entropy between the predicted probabilities and the correct output $\mathrm{CE}(f(x), y) :=  \log\left(\sum_{i} e^{f(x)_i}\right) - f(x)_y$, since it discourages interventions that inflate many logits simultaneously. 
We set a maximum number of 50 steps and re-run the algorithm with multiple perturbed starting values to encourage exploring different neuron sets. Details about the optimisation procedure can be found in Appendix \ref{sec:algo}.
Note that instead of directly estimating the interventions, we estimate $\mathbf{s}$, the scaling factor of the activation differences. This informs the optimiser about the expected changes in the activations, letting $\mathbf{a'}$ act as a preconditioner for the optimisation problem.
\begin{algorithm}
\caption{Compute DSI}
\label{alg:rel}
\begin{algorithmic}[1]\onehalfspacing
\small
\Require dataset $D^-$, dataset $D^+$, set size $n$, number of steps $t$
\State $\mathbf{a'} \gets \mathrm{mean}_{x \in D^+} (\mathbf{a}\vert x) - \mathrm{mean}_{x \in D^-} (\mathbf{a}\vert x)$
\Comment{average activation difference}
\State $\mathbf{g} \gets \mathrm{mean}_{x \in D^-} \; \nabla^r_\mathbf{a} \; f(x)_{y^+}$
\Comment{robustified gradient in 0-shot setting}
\State $\mathbf{e} \gets \mathbf{g} \odot \mathbf{a'}$
\Comment{expected first-order effect of interventions}
\State $s_0 \gets \mathbf{1}(\mathrm{topn}(\mathbf{e}))$
\Comment{most relevant neurons as starting point}
\For{$i = 1$ \textbf{to} $t$}
\State $s_i \approx \argmin_s \mathrm{mean}_{x \in D^-} \mathrm{CE}\left(f_{s \odot \mathbf{a'}}(x), y\right)$
\Comment{update intervention}
\\
\quad \quad \quad s.t. $\mathrm{nnz}(s) \leq n, \; s \text{ close to } s_{i-1}$
\Comment{sparse \& close to previous step}
\EndFor
\State \Return $s_t$
\end{algorithmic}
\end{algorithm}
\vspace{-4pt}

\section{Experiments} \label{sec:experiments}

\paragraph{Experimental setup} We evaluate our approach on three instruction-tuned language models from widely used families, Qwen 3 (8B) \citep{qwen3}, Gemma 3 (4B) \citep{gemma3}, and Llama 3.2 (3B) \citep{llama3} (see Appendix~\ref{app:models} for details), spanning a range of model depths (28-36 layers), attention heads (8-32 heads), and hidden dimensions (128, 256). We consider a subset of 12 tasks from~\citet{FV}, including two abstraction levels (see Table~\ref{tab:tasks} for an overview): \textit{extractive}, where the correct output is already present in the input (e.g., choosing the fruit from a set of words), and \textit{abstractive}, where the output is not present in the input (e.g., naming the capital of a given country).
For each task, we split the input-output pairs 50/50 into a training set and a test set. We then generate 100 0-shot prompts and 100 10-shot prompts from the training split. Two datasets contain fewer samples, resulting in smaller train and test splits. Details on the task data, splits, and prompt structure are provided in Appendix~\ref{ref:tasks}.
For the interventions, we consider the outputs of all attention heads in each layer at the last token position (before they are projected to the residual stream dimension).

\paragraph{Nonlinear, neuron-specific intervention effects} 
A key motivation for DSI is its ability to go beyond first-order effect estimates by accounting for nonlinear, neuron-specific effects and intervention effects across layers. Before proceeding to the main results, we illustrate the importance of this on the \textit{english–french} task using Llama 3.2 (3B) by studying interventions on the most relevant neurons identified via a first-order approximation using the robustified gradient, but without iterative refinement.
To this end, we randomly sample neurons and neuron pairs from the 16 neurons with the highest expected intervention effect and measure the logit of the correct token under single and pairwise interventions.

\begin{figure}[ht!]
    \centering
    \begin{subfigure}[b]{0.27\linewidth}
        \centering
        \includegraphics[width=\linewidth]{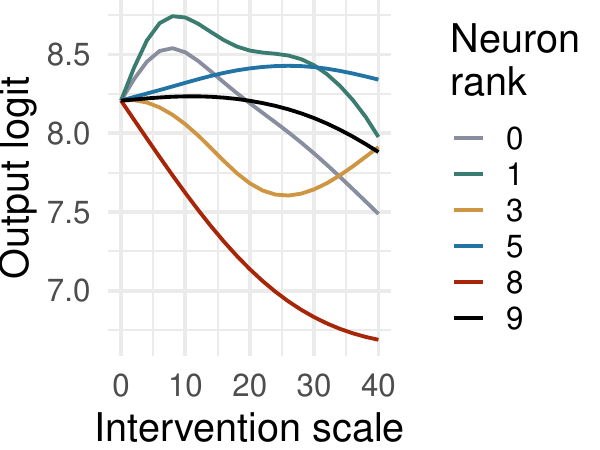}
        \caption{Neuron intervention effects}
        \label{fig:single}
    \end{subfigure}
    \hfill
    \begin{subfigure}[b]{0.71\linewidth}
        \centering
        \includegraphics[width=\linewidth]{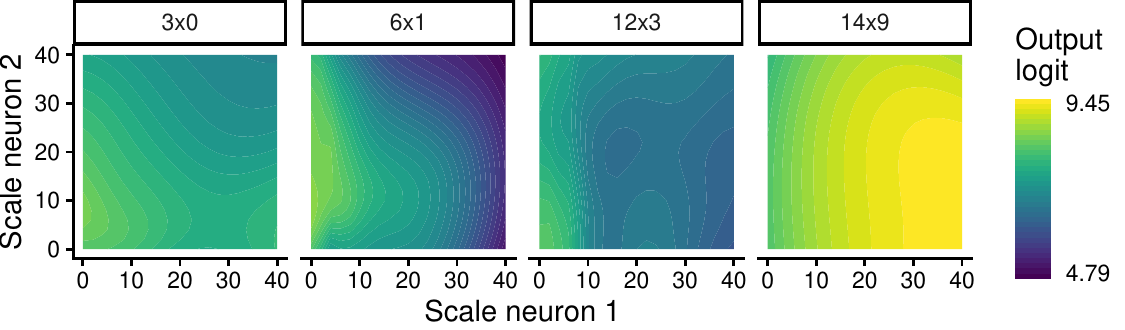}
        \caption{Pairwise neuron intervention effects}
        \label{fig:pairs}
    \end{subfigure}
    \caption{Value of the correct output logit of Llama 3.2 (3B) for the \textit{english-french} task, when intervening on hidden neurons. Neurons are sampled from the 16 neurons with the highest expected intervention effect, and labels correspond to their ranks. (a) shows the correct output logit as a function of single-neuron interventions, (b) as a function of interventions on pairs of neurons. Each pair consists of neurons from two different model layers.}
    \label{fig:int}
\end{figure}

Figure \ref{fig:single} shows the output logit corresponding to the correct token (averaged over $100$ 0-shot examples from the test set) for interventions on single neurons at a time.
The interventions are scaled by the average activation difference.
We observe that even among the neurons with the largest expected positive effect, some reduce the output logit, and that the logit varies nonlinearly with intervention strength. Figure \ref{fig:pairs} shows the effect of intervening on pairs of neurons. We observe interaction effects, such that the maximum logit corresponds to a setting where the intervention strength differs between neurons.
These observations indicate limitations of first-order effect estimation, as it does not reliably identify neurons with a positive effect. They also point to limitations of approaches that determine a uniform scaling factor for all neurons, as even after rescaling the intervention by the average activation difference, the optimal intervention strength varies substantially across neurons, motivating neuron-level, iterative techniques such as DSI.

\subsection{Evaluating DSI} \label{sec:eval}

For all tasks and models, we compute DSI for the neuron set sizes $n = 64$ and $n = d_\text{model}$ (the residual stream dimension, 2,560 to 4,096), and generate $200$ 0-shot prompts from the test split to evaluate the interventions. We limit ourselves to next-token prediction; if a correct output consists of multiple tokens, we evaluate whether the first predicted token is correct. We compare 0-shot task accuracy against the \textbf{baseline} model behaviour (no intervention) and \textbf{Function Vectors (FV)} \citep{FV}, an approach that constructs steering vectors in the residual stream by averaging the outputs of a set of attention heads with the largest effect on task execution across multiple tasks. We select FVs for comparison because of their widespread use, and use the tasks and datasets from the original paper.
Because FVs require an a priori decision regarding the intervention layer, two conditions are considered: one where the vector is added at $1/3$ of the model depth (as in \citet{FV}), and another where the best-performing layer per task is used.
Additionally, because the Gemma 3 model includes an extra normalisation layer after the attention block, FVs cannot be applied directly to this architecture. Therefore, we do not report FV results for this model.

\begin{table}[h]
\centering
\footnotesize
\caption{Average accuracy on the abstractive and extractive tasks in the 0-shot setting. Results for interventions on neuron set sizes $64$ and $d_\text{model}$ (the residual stream dimension of the model, 2,560 to 4,096). Baseline refers to model behaviour without interventions, "no rob." to no gradient robustification, "no iter." to no iterative refinement, "FV" to function vectors,  "FV best" to FVs evaluated at the best-performing layer. Ranges refer to the minimum and maximum task accuracy. Best-performing methods are \textbf{bold}, second-best \underline{underlined}. Across almost all conditions, DSI with gradient robustification and iterative refinement performs best.}
\vspace{-6pt}
\label{table:main}
\begin{tabular}{rrrrrrrr}
\toprule
\multicolumn{2}{c}{ } & \multicolumn{3}{c}{abstractive} & \multicolumn{3}{c}{extractive} \\
\cmidrule(l{3pt}r{3pt}){3-5} \cmidrule(l{3pt}r{3pt}){6-8}
size & method & Llama 3.2  & Qwen 3  & Gemma 3  & Llama 3.2   & Qwen 3   & Gemma 3  \\
\midrule
$0$ & baseline & 0.02 {\tiny0.00-0.06} & 0.04 {\tiny0.00-0.12} & 0.01 {\tiny0.00-0.03} & 0.28 {\tiny0.14-0.52} & 0.03 {\tiny0.01-0.07} & 0.12 {\tiny0.03-0.27}\\
$d_\text{model}$ & FV & 0.36 {\tiny0.00-0.64} & 0.16 {\tiny0.00-0.86} & / & 0.43 {\tiny0.22-0.96} & 0.41 {\tiny0.04-0.97} & / \\
$d_\text{model}$ & FV best & 0.57 {\tiny0.00-0.92} & 0.33 {\tiny0.00-0.97} & / & 0.63 {\tiny0.36-0.97} & 0.43 {\tiny0.17-0.88} & / \\
\midrule
64 & no rob. & 0.65 {\tiny0.17-0.98} & 0.59 {\tiny0.04-0.97} & 0.74 {\tiny0.63-1.00} & 0.56 {\tiny0.33-0.93} & 0.51 {\tiny0.12-0.95} & 0.78 {\tiny0.42-1.00}\\
64 & no iter. & 0.66 {\tiny0.27-0.97} & 0.53 {\tiny0.09-0.94} & 0.72 {\tiny0.57-1.00} & 0.73 {\tiny0.26-0.99} & 0.65 {\tiny0.35-1.00} & 0.74 {\tiny0.35-1.00}\\
64 & DSI & \underline{0.89} {\tiny0.74-0.98} & \underline{0.87} {\tiny0.70-0.98} & 0.86 {\tiny0.70-1.00} & \underline{0.98} {\tiny0.95-1.00} & \underline{0.98} {\tiny0.97-1.00} & \underline{0.96} {\tiny0.94-0.99}\\
\midrule
$d_\text{model}$ & no rob. & 0.85 {\tiny0.69-0.94} & 0.60 {\tiny0.10-0.93} & 0.85 {\tiny0.73-0.99} & 0.86 {\tiny0.46-0.98} & 0.81 {\tiny0.48-0.99} & 0.80 {\tiny0.31-1.00}\\
$d_\text{model}$ & no iter. & 0.84 {\tiny0.64-0.99} & 0.80 {\tiny0.67-0.95} & \underline{0.88} {\tiny0.76-0.99} & 0.85 {\tiny0.32-0.99} & 0.83 {\tiny0.45-1.00} & 0.85 {\tiny0.35-1.00}\\
$d_\text{model}$ & DSI & \textbf{0.90} {\tiny0.73-0.98} &\textbf{0.89} {\tiny0.67-0.98} & \textbf{0.89} {\tiny0.74-1.00} & \textbf{0.99} {\tiny0.98-1.00} & \textbf{0.99} {\tiny0.98-1.00} & \textbf{0.99} {\tiny0.98-1.00}\\
\bottomrule
\end{tabular}
\end{table}

We also compare against two additional conditions evaluating iterative refinement and gradient modifications:
(1) \textbf{no iterative refinement}: the $n$ interventions with the largest effects as approximated via first-order Taylor expansion are used (i.e., the starting values in Algorithm \ref{alg:rel}), and one joint scaling factor $s \in [0.01, 50]$ is determined for all interventions.
(2) \textbf{no gradient robustification}: the same as (1), but in addition, the first-order effects are approximated without the gradient modifications described in Section \ref{sec:lrp}.

\paragraph{Main results} Table \ref{table:main} shows the resulting accuracy from applying DSI in the 0-shot setting across models, averaged over the two task categories (abstractive and extractive, 6 tasks each), for the intervention set sizes $64$ and $d_\text{model}$. Across model families, DSI substantially improves task accuracy over the no-intervention and FV baselines, with interventions on $64$ neurons achieving average accuracies ranging from 0.86 to 0.98 across settings. We observe that the accuracy on $64$ neurons almost matches the full model dimension, indicating that sparse interventions can be effective.
The results further indicate that both gradient robustification and iterative refinement contribute consistently to improved intervention performance.

\subsection{Sparse Interventions} \label{sec:sparse}

\begin{figure}[h]
{\centering
    \includegraphics[width=\textwidth]{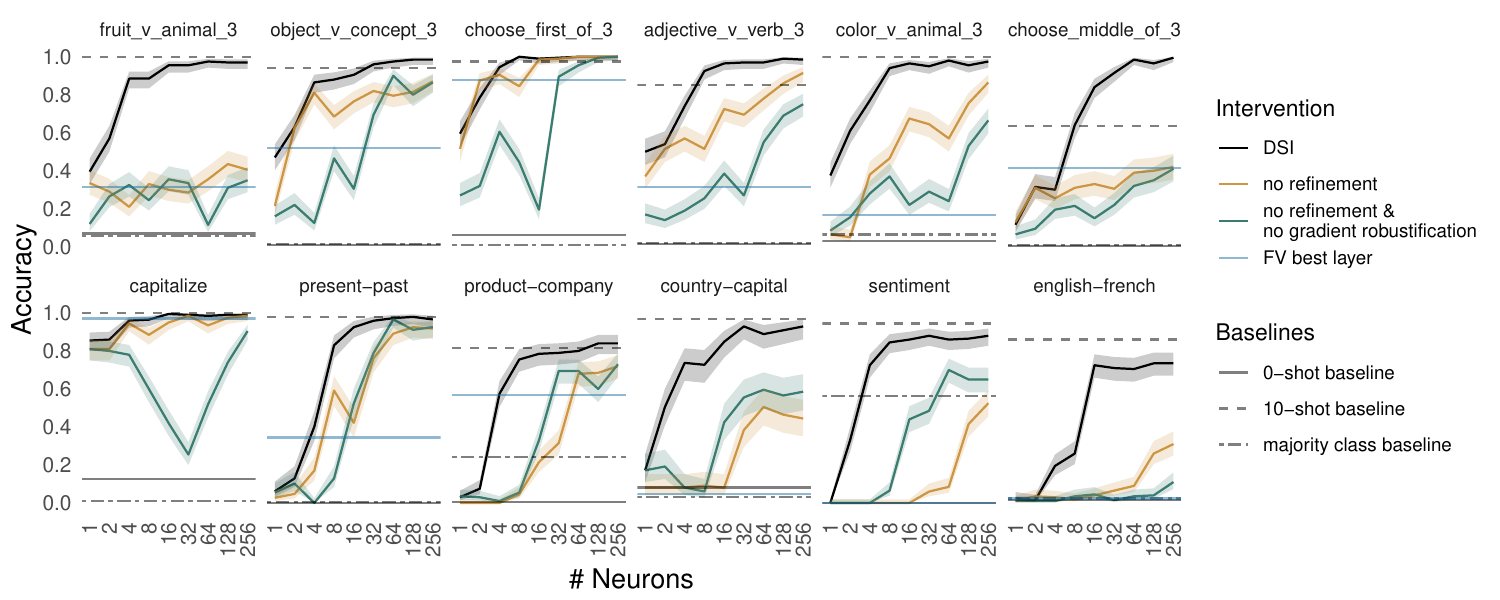}}
\vspace{-10pt}
\caption{0-shot test accuracy for interventions across set sizes for extractive (top row) and abstractive (bottom row) tasks. Results shown are for Qwen 3 (8B); additional results across models are provided in Appendix~\ref{app:results}. Shaded regions correspond to 95\% confidence intervals. 0-shot and 10-shot baselines correspond to ICL performance without intervention; the majority-class baseline to the proportion of the most common correct outputs, and FV best layer to Function Vectors at the best-performing layer.} \label{fig:acc}
\end{figure}

Figure~\ref{fig:acc} extends this analysis by showing accuracy across all 12 tasks as a function of set sizes from 1 to 256 for Qwen 3 (8B), including conditions without gradient robustification and without iterative refinement. Additional results for other models are provided in Appendix~\ref{app:results}. Shaded regions correspond to 95\% Wilson confidence intervals \citep{wilsonci}, taking into account uncertainty due to the limited number of test samples. Across settings, particularly for small set sizes, DSI consistently achieves the highest task accuracy, and both iterative refinement and gradient robustification are necessary for high performance, with removal of either degrading task accuracy. These results overall demonstrate that across a range of tasks, interventions on as few as 4--32 neurons activate task-solving behaviour, highlighting DSI's potential for identifying sparse and effective neuron-level interventions.

\subsection{Distributed Interventions} \label{sec:distributed}
\begin{wrapfigure}[10]{r}{0.4\textwidth}
\vspace{-20pt}
\includegraphics[width=.38\textwidth]{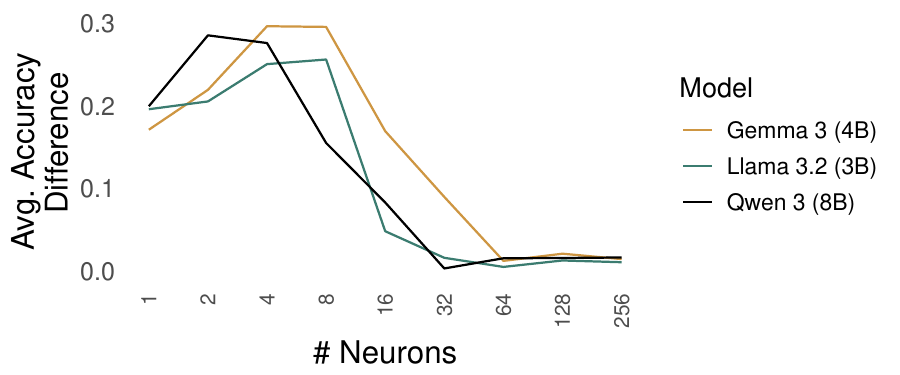}
\caption{Average accuracy difference
between single-layer and distributed DSI
interventions across all 12 tasks} \label{fig:sl}
\vspace{-1em}
\end{wrapfigure}

To assess the relative effectiveness of distributed versus layer-localized interventions, we applied DSI restricted to a single layer at one-third of model depth, following \citet{FV}. Figure~\ref{fig:sl} presents the resulting accuracy differences between single-layer and distributed interventions, averaged across all extractive and abstractive tasks. 
For neuron set sizes of 64 or greater, effective interventions can be achieved within a single layer. For smaller set sizes, distributed interventions provide a substantial improvement in accuracy. These results suggest that the design of sparse interventions requires methods that can simultaneously estimate interventions across multiple layers. Appendix~\ref{sec:pos} presents further results on the distribution of identified neurons across attention heads. For instance, at set size 64, the selected neurons span a variety of attention heads across layers, primarily in middle and later layers, highlighting that neuron selection is distributed across the model and varies across tasks and models.

\section{Identifying distributed concept representations} \label{sec:axbench}

\paragraph{Open-ended concept steering} Steering open-ended generation presents a challenging setting for model interventions: methods must alter specific properties of a response while preserving instruction following and fluency. To generalise our evaluations of DSI to autoregressive settings, we examine how the incorporation of concepts characterised by short textual descriptions can be achieved through distributed neural computation. We apply DSI to AxBench~\citep{wu2025axbench}, a standardised benchmark evaluating how well steering methods incorporate concepts during prompt answering.
Each concept corresponds to a sparse autoencoder feature extracted from Gemma-2~\citep{gemmateam2024gemma2improvingopen, gemmascope}. We apply DSI to concepts extracted at layer 20 for two model sizes (2B and 9B), with 500 concepts per model size.
For each concept, the dataset includes 72 positive examples generated with GPT-4o-mini instructed to include the concept and 72 negative examples from model outputs without intervention.
For \textbf{evaluation}, we follow the AxBench setup, including their evaluation prompts and templates. An LLM judge\footnote{Version \texttt{gpt-4o-mini-2024-07-18}, accessed via the OpenAI API.} evaluates responses to 10 randomly sampled instructions from AlpacaEval~\citep{alpaca_eval}, using 5 of the instructions to calibrate the strength of the steering interventions, and the remaining 5 to evaluate steering performance.
Each response is rated as the harmonic mean of three dimensions: (1) \textit{Concept}: how strongly is the concept present in the response, (2) \textit{Instruct}: how well is the prompt's instruction followed, and (3) \textit{Fluency}: how fluent is the response. The scores on each dimension range from 0 to 2; if any of the dimensions is rated as  0, the overall steering score is set to 0. Details on adapting DSI to the generative setting can be found in Appendix~\ref{app:axbench}.

\begin{figure}[th]
 \centering
{\centering
    \includegraphics[width=\textwidth]{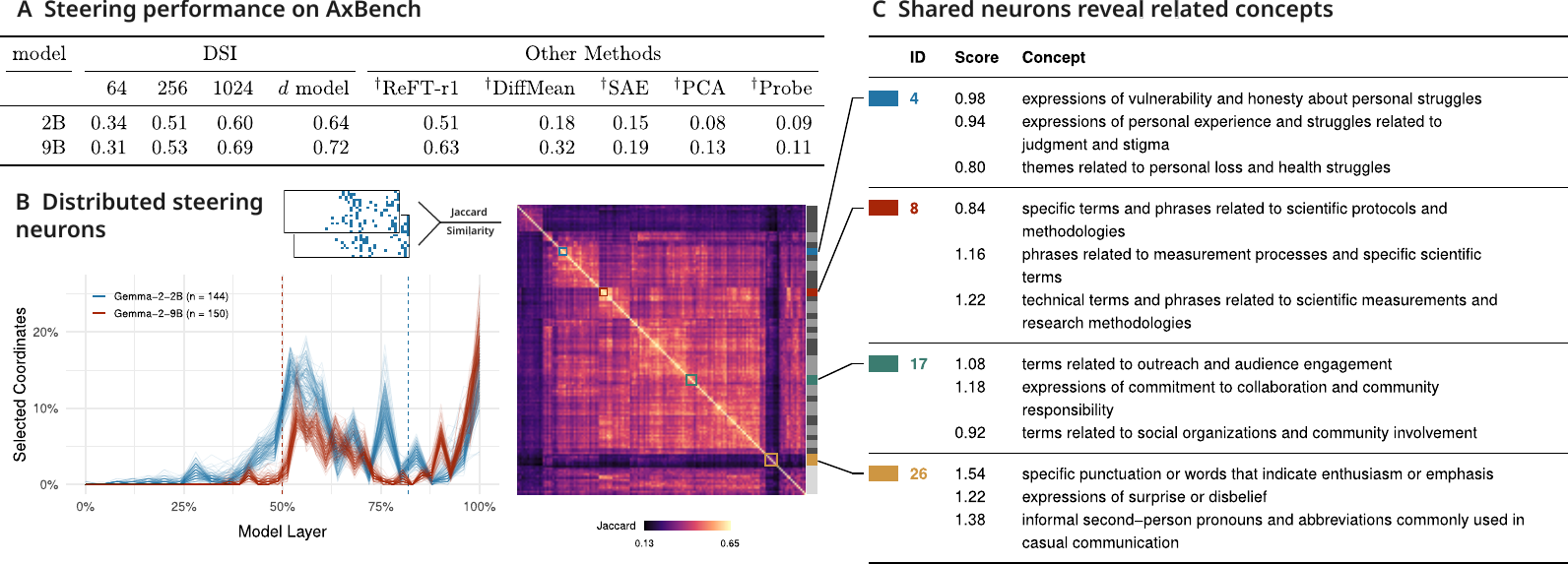}}
\vspace{-10pt}
\caption{DSI interventions on AxBench concepts. \textbf{A} Steering performance of DSI on AxBench for different set sizes, $d$~model (the residual stream of the model) is equal to 2304/3584 neurons for the 2B/9B parameter model. $^\dagger$From \citet{wu2025axbench}, Table 2. \textbf{B} Distribution of selected neuron coordinates across layers for DSI interventions on 256 neurons. Each line corresponds to a text concept; concepts are filtered for an AxBench score above 0.75. Dashed lines signify layer 20; the concept extraction layer. \textbf{C} Concepts are clustered by the Jaccard similarity of the intervention sets. Shown are example concepts with the highest average similarity within the cluster; "Score" corresponds to the AxBench steering score.} \label{fig:clustering}
\vspace{-3pt}
\end{figure}

\paragraph{Effective generative steering with DSI} 
We evaluate DSI across neuron set sizes from 64 to the full model dimension (2304/3584 for Gemma-2 2B/9B). As shown in Figure~\ref{fig:clustering} (A), DSI achieves similar performance to recent dense steering methods when matching the number of neurons to the latent model dimensionality $d$.
Limiting the neuron set size to 1024, roughly half the neuron budget for the 2B model and one third for the 9B model, maintains steering performance and demonstrates that effective steering can be achieved with a substantially reduced intervention footprint.
Reducing the neuron set further reveals a trade-off between steering performance and sparsity, with DSI using 64 neurons still achieving performance on par with dense steering baselines. 

\paragraph{Layerwise distribution of steering neurons}
Next, we focus on interventions of size 256, balancing sparsity and performance, on the 9B model.
To study concept steering in natural language generation, we focus on the text genre and retain 150 concepts with an  overall steering score above 0.75, ensuring that the corresponding interventions successfully induce the target concept. Figure~\ref{fig:clustering} (B) shows the distribution of intervened-on neurons across layers for these concepts, highlighting that most steering neurons are located in late mid to late layers, with almost no interventions in the first third of the model. This pattern is remarkably consistent across concepts: we identify 81 neurons that are present in more than 80\% of the interventions, despite the concepts spanning diverse categories, including grammatical and syntactic patterns, semantic descriptions, and pragmatics (see Table \ref{tab:clusters_9b} in the Appendix). This suggests that these neurons support shared computational processes that coordinate concept expression in text generation, mirroring related findings for task switching at the level of attention heads~\citep{FV}.

\paragraph{Steering neurons capture concept relationships} 
We next examine whether concepts with higher overlap in their intervened-upon neuron sets tend to be semantically related. We quantify the overlap between intervention sets using the Jaccard index, defined as the number of shared neurons divided by the total number of unique neurons in the two sets. After performing hierarchical agglomerative clustering, we identify 26 clusters covering 135 concepts, with 15 concepts left unassigned (see Appendix \ref{app:clustering} for details on the clustering procedure, a list of all identified clusters and concepts and the results on the 2B model). We observe that concepts group into diverse clusters, with most clusters exhibiting high semantic coherence. Figure \ref{fig:clustering} (C) providing examples across topics such as personal struggles, science, or informal speech. This indicates that neuron sets identified solely by their ability to induce a target concept also capture meaningful relationships across concepts, providing a link between observable model behaviour and its underlying computation. 

Overall, these results demonstrate that DSI identifies sparse and distributed interventions that retain strong performance for steering open-ended text generation. The concentration of interventions in specific layers and sharing of neurons across interventions suggests that concept steering relies heavily on a relatively small set of model components. Moreover, the semantic coherence of clusters derived solely from intervention-set overlap indicates that concepts with similar intervention signatures tend to be semantically related, suggesting that the selected model neurons reflect concept-specific structure.

\section{Discussion}

We introduced \textit{Distributed Sparse Interventions (DSI)}, a novel steering method for identifying neuron-level interventions.
By operating directly on individual neurons rather than latent-space vectors, DSI can effectively capture interactions and nonlinear effects between neurons while providing fine-grained control over intervention strength across the model.
We demonstrated that DSI enables successful steering in both abstractive and extractive task settings using surprisingly small sets of 4--32 neurons distributed across model layers, and extends effectively to concept steering in generative settings.

While DSI selects intervention neurons solely based on their ability to induce the desired target behaviour, we find that greater neuron-level overlap between intervention sets corresponds to greater functional relatedness, suggesting that neuron-level sets reveal a correspondence between observable model behaviour and underlying computation. Thus, although task-relevant neurons are distributed across the model, their participation in steering appears structured rather than arbitrary. 

By modelling nonlinear causal effects in steering vectors, \cite{bigoulaeva-etal-2026-patches} recently demonstrated that localised interventions at specific layers are involved in modulating task circuits. Our results provide further evidence for localised information processing at the neuron level. Characterising neuron-level computations may enable more precise and robust steering by targeting specific computational components rather than inducing global shifts with potentially entangled downstream effects, while also making interventions more interpretable and amenable to safeguarding.

\paragraph{Limitations}
Our evaluation covers a selected set of models and tasks, and the effectiveness of the identified intervention patterns across other architectures and steering settings requires further evaluation. For open-ended generation, future work could explore combining DSI and token-dependent intervention strengths or alternative objectives, such as preference-based losses. Finally, DSI is computationally more intensive than non-iterative steering methods.

\bibliography{references}
\bibliographystyle{iclr2027_conference}

\appendix

\newpage

\section{Interactions between interventions} \label{app:toy}

Several components in transformer models can induce interactions between interventions. For example, products (the self-attention mechanism) induce interactions between interventions on the multiplied neurons, while nonlinearities (activation functions) induce interactions between interventions on neurons connected across layers. Here, we study an example of how product terms induce such interactions, and how an iterative computation of interventions can better account for them.

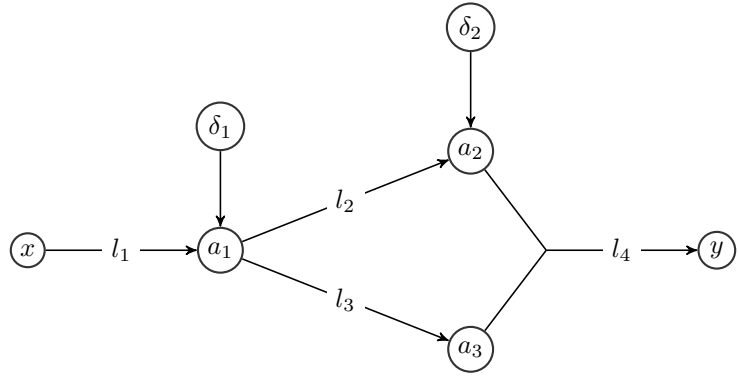
\begin{figure}[H]
    \begin{center}
    \vspace{.5cm}
    \begin{tikzpicture}[latent/.style={circle,draw=black!80,thick,inner sep=2,text centered}]

    \begin{scope}

    \node[latent](x){$a_1$};
    \coordinate[right=3 of x] (xhelp) {};
    \coordinate[right=1 of xhelp] (hhelp) {};

    \node[latent, node distance = 2](xx)[left=of x]{$x$};
    \draw [->,>=stealth',semithick] (xx) -- node [fill=white] {$l_1$} (x);

    \node[latent, node distance = 1](y1)[above=of xhelp]{$a_2$};
    \node[latent, node distance = 1](y2)[below=of xhelp]{$a_3$};

    \node[latent, node distance = 3](z)[right=of xhelp]{$y$};

    \node[latent, node distance = 1](epsilon)[above=of y1]{$\delta_2$};
    \draw [->,>=stealth',semithick] (epsilon) -- (y1);

    \node[latent, node distance = 1](delta)[above=of x]{$\delta_1$};
    \draw [->,>=stealth',semithick] (delta) -- (x);

    \draw [->,>=stealth',semithick] (x) -- node [fill=white] {$l_2$} (y1);
    \draw [->,>=stealth',semithick] (x) -- node [fill=white] {$l_3$} (y2);
    \draw [-,>=stealth',semithick] (y1) -- (hhelp);
    \draw [-,>=stealth',semithick] (y2) -- (hhelp);
    \draw [->,>=stealth',semithick] (hhelp) -- node [fill=white] {$l_4$} (z);

    \end{scope}

    \end{tikzpicture}
    \end{center}
    \caption{Example system with causal interventions $\delta_1$ and $\delta_2$.} \label{fig:example}
\end{figure}

Consider the model $f: \R \to \R$, $x \mapsto l_4(l_2(l_1(x)), l_3(l_1(x)))$ depicted in Figure \ref{fig:example}. Setting $l_1 = l_2 = l_3 = \mathrm{id}$ and $l_4(a_2, a_3) = a_2a_3$, it reduces to $f(x) = x^2$.
The effects of $\delta_1$ and $\delta_2$ in isolation are

\begin{align}
\Eff(\delta_1) &= f(x|a_1 \gets a_1 + \delta_1) - f(x) = (x+\delta_1)^2 - x^2 = 2x\delta_1 + \delta_1^2 \\
\Eff(\delta_2) &= f(x|a_2 \gets a_2 + \delta_2) - f(x) = (a_2 + \delta_2)a_3 - a_2a_3 = x\delta_2
\end{align}

Now, consider adding both effects simultaneously:

\begin{align}
\Eff(\delta_1, \delta_2) &= f(x | a_1 \gets a_1 + \delta_1, a_2 \gets a_2 + \delta_2) - f(x) \\
&= 2x\delta_1 + \delta_1^2 + x\delta_2 + \delta_1 \delta_2 \\
&= \Eff(\delta_1) + \Eff(\delta_2) + \delta_1 \delta_2.
\end{align}

The effect of both interventions combined is not equal to the sum of the intervention effects in isolation, since an interaction term $\delta_1\delta_2$ occurs.
This interaction term is therefore also absent from the first-order Taylor approximation of the intervention effects:

\begin{align}
\bar \Eff(\delta_1) &= \partial_{a_1} f(x) \delta_1 = 2a_1\delta_1 = 2x\delta_1 \\
\bar \Eff(\delta_2) &= \partial_{a_2} f(x) \delta_2 = a_3\delta_2 = x\delta_2
\end{align}

However, this effect can be mitigated through an iterative approach. Consider the setting where the effects have been approximated and the intervention with the highest effect (e.g., $\delta_2$) has been added to the model. Now, to add more interventions, we re-approximate at the new root point, that is, we approximate $f_{\delta_2}(x) = f(x|a_2 \gets a_2 + \delta_2)$:

\begin{equation}
    \bar \Eff(\delta_1) = \partial_{a_1} f_{\delta_2}(x) \delta_1 = (2a_1 + \delta_2)\delta_1 = 2x\delta_1 + \delta_1 \delta_2
\end{equation}

and we can see that at the new root point, the approximation becomes more precise as it includes the interaction effect between the interventions.

\section{Results for LLama 3.2 and Gemma 3} \label{app:results}

\begin{figure}[H]
{\centering
    \includegraphics[width=\textwidth]{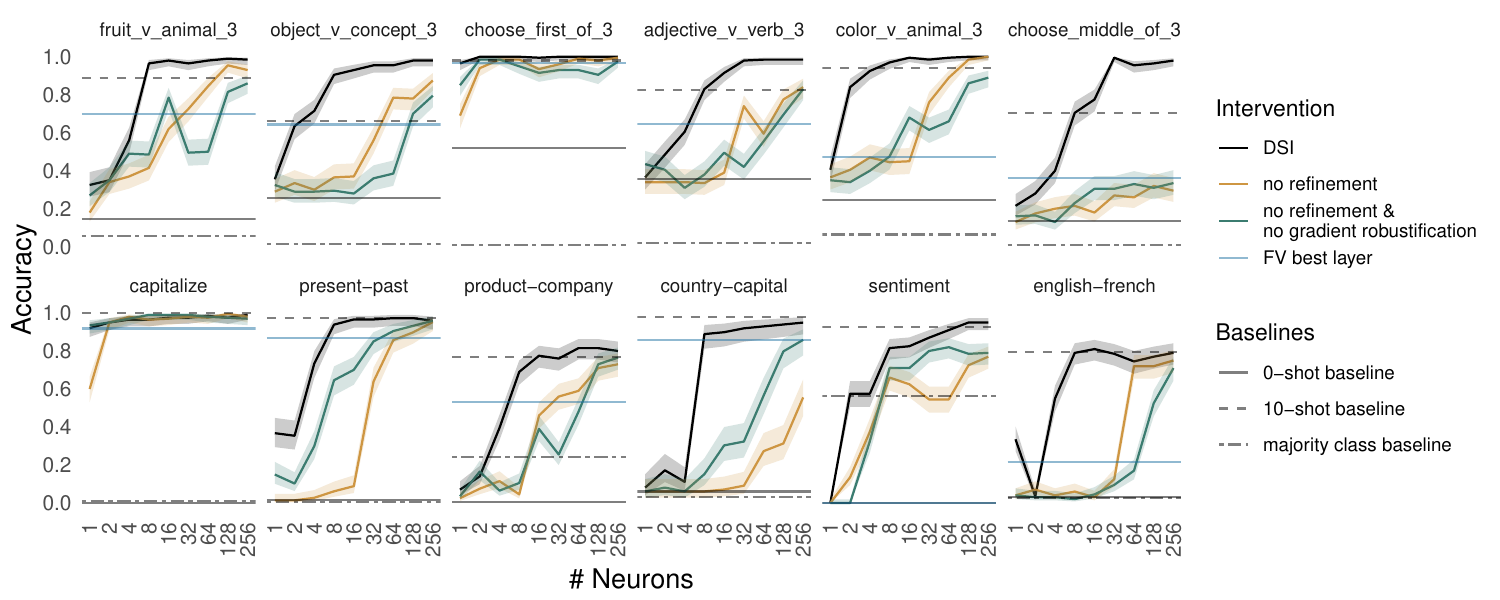}
}
\caption{0-shot test set accuracy for interventions across set sizes for extractive (top row) and abstractive (bottom row) tasks. Results shown are for Llama 3.2 (3B), shaded regions correspond to 95 \% confidence intervals. 0-shot and 10-shot baselines correspond to ICL performance without intervention, majority class baseline to the proportion of the most common correct output token, and FV best layer to the function vector intervention added at the best performing layer.} \label{fig:llama}
\end{figure}

\begin{figure}[H]
{\centering
    \includegraphics[width=\textwidth]{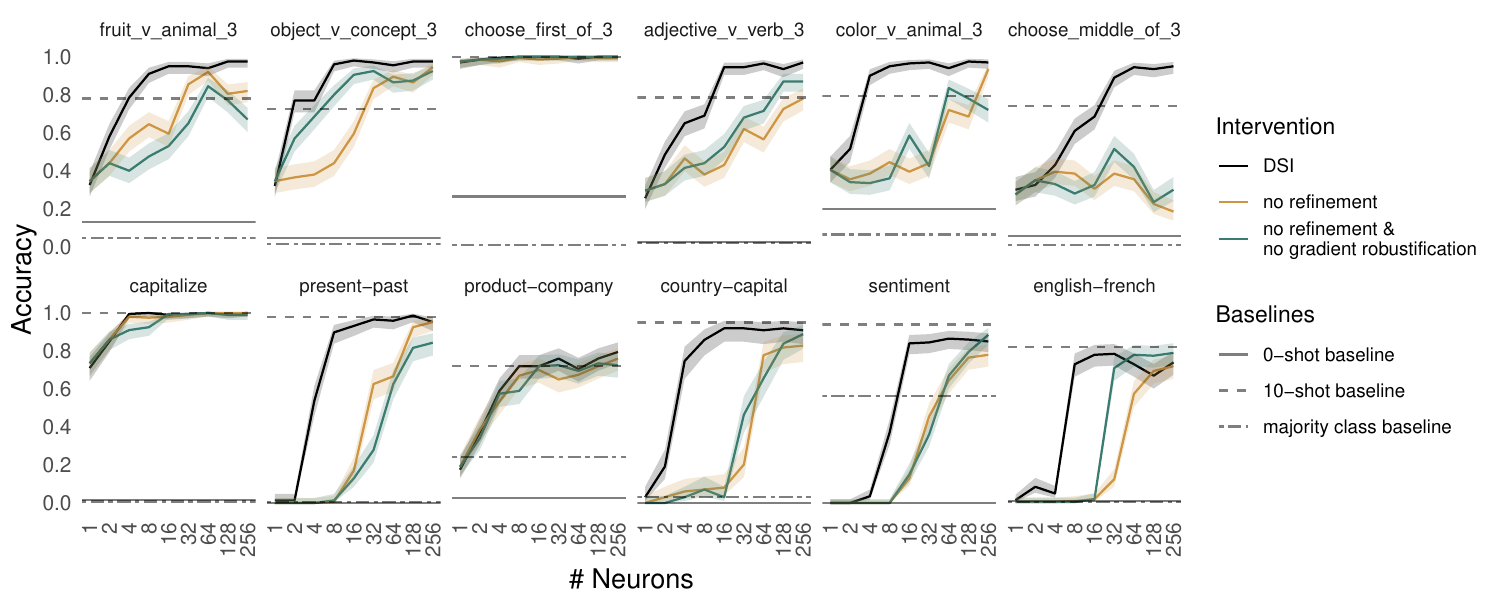}
}
\vspace{-8pt}
\caption{0-shot test set accuracy for interventions across set sizes for extractive (top row) and abstractive (bottom row) tasks. Results shown are for Gemma 3 (4B), shaded regions correspond to 95 \% confidence intervals. 0-shot and 10-shot baselines correspond to ICL performance without intervention, majority class baseline to the proportion of the most common correct output token, and FV best layer to the function vector intervention added at the best performing layer.} \label{fig:gemma}
\end{figure}

Figures \ref{fig:llama} and \ref{fig:gemma} show the accuracy by intervention set size for the LLama 3.2 (3B) and Gemma 3 (4B) models, respectively. These figures are analogous to Figure~\ref{fig:acc} in the main text.

\section{Model details} \label{app:models}
For all models, we use the model implementations provided by TransformerLens \citep{transformerlens}. TransformerLens translates the model implementation into a standardised set of layers and centers the unembedding. Table \ref{table:models} provides the corresponding TransformerLens identifiers as well as the model dimensions.

\begin{table}[h]
\centering
\scriptsize
\caption{Properties of the studied models. The total number of considered neurons for the DSI intervention is the product of model layers, heads, and the number of neurons per head.}
\vspace{1ex}
\begin{tabular}{rrrrrr}
\toprule
Model & TransformerLens ID & \# Layers & \# Heads & \# Neurons p.h. & \# Neurons total\\
\midrule
Qwen 3 (8B) & \texttt{Qwen/Qwen3-8B} & 36 & 32 & 128 & \num{147456}\\
Gemma 3 (4B) & \texttt{google/gemma-3-4b-it} & 34 & 8 & 256 & \num{69632}\\
Llama 3.2 (3B) & \texttt{meta-llama/Llama-3.2-3B-Instruct} & 28 & 24 & 128 & \num{86016}\\
\bottomrule
\end{tabular}
\vspace{1em}
\label{table:models}
\end{table}

\section{Tasks details} \label{ref:tasks}

Table \ref{tab:tasks} provides an overview of the tasks used in Section~\ref{sec:experiments}. We use all tasks as provided in the supplementary material to \citet{FV}, accessible at the accompanying GitHub repository\footnote{\url{https://github.com/ericwtodd/function_vectors} (commit fb9eac7b6dc707ea1475a717379916007fe448d5)} under MIT License. We split all tasks 50/50 into a training and a test set. To generate $k$-shot ICL examples, we follow the default template of \citet{FV} and present $k$ input-output examples follow by an example to complete as "Q: input$_1$\textbackslash{}nA: output$_1$\textbackslash{}n\textbackslash{}nQ: input$_2$\textbackslash{}nA: output$_2$\textbackslash{}n\textbackslash{}n[...]Q: input$_k$\textbackslash{}nA: output$_k$\textbackslash{}n\textbackslash{}nQ: input\textbackslash{}nA:". 0-shot examples are of the form "Q: input\textbackslash{}nA:". From the training set, we generate $100$ 0-shot and $100$ 10-shot prompts to compute interventions. From the test set, we generate $200$ 0-shot prompts to compute the reported accuracies. All tasks contain at least $200$ test set samples, except for \textbf{present-past} (146 training set and 147 test set examples) and \textbf{country-capital} (98 training and 99 test set examples) - these tasks have been trained and evaluated on fewer examples, respectively, and their confidence intervals reflect the higher uncertainty level.

\begin{table}[h]
\caption{Overview of the 12 tasks used in the main experiments in Section~\ref{sec:experiments}.}
\centering
\small
\vspace{1ex}
\begin{tabular}{lp{5cm}p{5cm}}
\toprule
\textbf{Task} & \textbf{Description} & \textbf{Example} \\
\midrule
\texttt{capitalize} & Capitalize the first letter of a word. & \textit{curtain} $\to$ \textit{Curtain} \\
\addlinespace
\texttt{country-capital} & Given a country, output its capital city. & \textit{Ecuador} $\to$ \textit{Quito} \\
\addlinespace
\texttt{english-french} & Translate an English word into French. & \textit{loved} $\to$ \textit{aimé} \\
\addlinespace
\texttt{present-past} & Convert a verb from present to past tense. & \textit{teach} $\to$ \textit{taught} \\
\addlinespace
\texttt{sentiment} & Classify the sentiment of a short text as positive or negative. & \textit{``Beautifully produced.''} $\to$ \textit{positive} \\
\addlinespace
\texttt{product-company} & Given a product, identify the company that created it. & \textit{Office Open XML} $\to$ \textit{Microsoft} \\
\addlinespace
\texttt{fruit\_v\_animal\_3} & Given two animals and a fruit, identify the fruit. & \textit{date, penguin, reindeer} $\to$ \textit{date} \\
\addlinespace
\texttt{color\_v\_animal\_3} & Given two animals and a color, identify the color. & \textit{eel, olive, turtle} $\to$ \textit{olive} \\
\addlinespace
\texttt{object\_v\_concept\_3} & Given two abstract concepts and one physical object, identify the object. & \textit{about, think, donut} $\to$ \textit{donut} \\
\addlinespace
\texttt{choose\_first\_of\_3} & Given three words, output the first one. & \textit{zealous, kindhearted, plant} $\to$ \textit{zealous} \\
\addlinespace
\texttt{adjective\_v\_verb\_3} & Given two verbs and one adjective, identify the adjective. & \textit{search, thirsty, manipulate} $\to$ \textit{thirsty} \\
\addlinespace
\texttt{choose\_middle\_of\_3} & Given three words, output the middle one. & \textit{quokka, hot, sail} $\to$ \textit{hot} \\
\bottomrule
\vspace{.5em}
\end{tabular}
\label{tab:tasks}
\end{table}

\citet{FV} adapted some of the tasks from earlier work: the \textbf{english-french} translation task is an adaptation from \citet{englishfrench}, the \textbf{product-company} task is adapted from \citet{productcompany}, and the sentiment analysis dataset is an adaptation from \citet{sentiment1} and \citet{sentiment2}.

For Section \ref{sec:iterative}, we generate a dataset for a \textbf{copy} task by using the inputs of the \textbf{english-french} task, and set the output equal to the input (i.e., the datapoint \textit{loved $\to$ aimé} becomes \textit{loved $\to$ loved}).

\section{LRP-modified gradient computation} \label{app:lrp}
To implement the modified gradients, we adapt the respective PyTorch modules implemented in TransformerLens \citep{transformerlens}. The modifications \texttt{detach} parts of the computation graph, leaving the forward pass unchanged, but alter the backward pass behaviour. Table \ref{tab:lrp-rules} contains an overview of the applied rules and how they are implemented.

\begin{table}[h]
\caption{LRP rules applied to transformer modules. $\mathrm{sg}(\cdot)$ denotes the stop-gradient (detach) operator.} \label{tab:lrp-rules}
\vspace{1ex}
\centering
\begin{tabular}{lll}
\toprule
\textbf{Module} & \textbf{Rule} & \textbf{Modified Forward Pass} \\
\midrule
Activation function $g(a)$ & Identity rule &
  $\tilde{a} = a \cdot \mathrm{sg}\!\left(\dfrac{g(a)}{a}\right)$ \\[10pt]
Gated MLP with activations $a$ & Half rule &
  $\tilde{a} = \tfrac{1}{2}\,a + \tfrac{1}{2}\,\mathrm{sg}(a)$ \\[6pt]
Attention with $\alpha$/V matrices $M$ & Half rule &
  $\tilde{M} = \tfrac{1}{2}\,M + \tfrac{1}{2}\,\mathrm{sg}(M)$ \\[6pt]
RMSNormPre with activation vector $\mathbf{a}$ & LayerNorm rule &
  $\mathbf{\tilde{a}} = \dfrac{\mathbf{a}}{\mathrm{sg}(\sigma)}, \quad
   \sigma = \sqrt{\dfrac{1}{d}\textstyle\sum_i \mathbf{a}_i^2 + \varepsilon}$ \\[10pt]
\bottomrule
\end{tabular}
\end{table}

For nonlinear activation functions $g$ that act on individual (scalar) activations at a time (e.g., SiLU), we employ the \textbf{Identity rule} \citep{NEURIPS2024_d6d0e41e} which treats $g$ as constant, preserving the magnitude of the effect estimate.
Gated MLPs and the attention mechanism contain multiplications of activations. In these settings, we employ the \textbf{Half rule} \citep{lrplstm, attnlrp} which prevents the effect estimate from doubling during the backward pass.
For layer normalization implemented by RMSNormPre modules, we use the \textbf{LayerNorm rule}, as described in \citep{transformerlrp}, to avoid an effect estimation collapse.

\section{Algorithm details} \label{sec:algo}

For the iterative refinement of the interventions, we use ZeroFPR \citep{zerofpr}, a numerical optimisation algorithm suited for the sum of a non-convex differentiable function and a nonsmooth penalty term.
To ensure sparsity of the identified neuron sets, we use the $L_0$ pseudo-norm ball as a penalty.
\begin{align*}
\iota_{B_0(n)}(s) \coloneqq \begin{cases} 0 &\mathrm{for } \; \mathrm{nnz}(s) <= n \\ \infty &\mathrm{for } \; \mathrm{nnz}(s) > n \end{cases}
\end{align*}
Here, $\mathrm{nnz}(s)$ corresponds to the number of non-zero elements in $s$ and $n$ is the desired neuron set size. In our implementation of the algorithm, we follow the implementation in ProximalAlgorithms.jl \citep{proxalg}.
Specifically, we use a convergence tolerance of $10^{-8}$ with a maximum number of 50 steps, adaptive step size estimation, a maximum number of $20$ line search backtracks, and L-BFGS \citep{lbfgs} to compute the line-search directions with a memory of 5 steps.
To facilitate escaping local minima near the starting values, we run the algorithm 10 times with perturbed starting values.
For the first run, we start with $s_0^1 = \mathbf{1}(\mathrm{topn}(\mathbf{e}))$, i.e., we set all parameters to $0$ except for the top $n$ largest effects approximated with a first-order approximation (as described in Section \ref{sec:initial}).
For successive runs, we start with random noise $s_0^i \sim \mathcal{N}(0, 0.3)$. We then take an initial step $\mathrm{topn}(s_0^i + 1.5 \; \gamma \nabla f(s_0^i))$ where $\gamma$ is the initial step size guess of ZeroFPR. This allows to sample different gradients at points close to $s_0^1$, while a factor of $1.5$ applied to the initial step size encourages the algorithm to explore regions beyond the limits of the algorithm's usual first step. Successive steps are taken without step size rescaling.

\begin{figure}[H]
    \centering
    \includegraphics[width=1\linewidth]{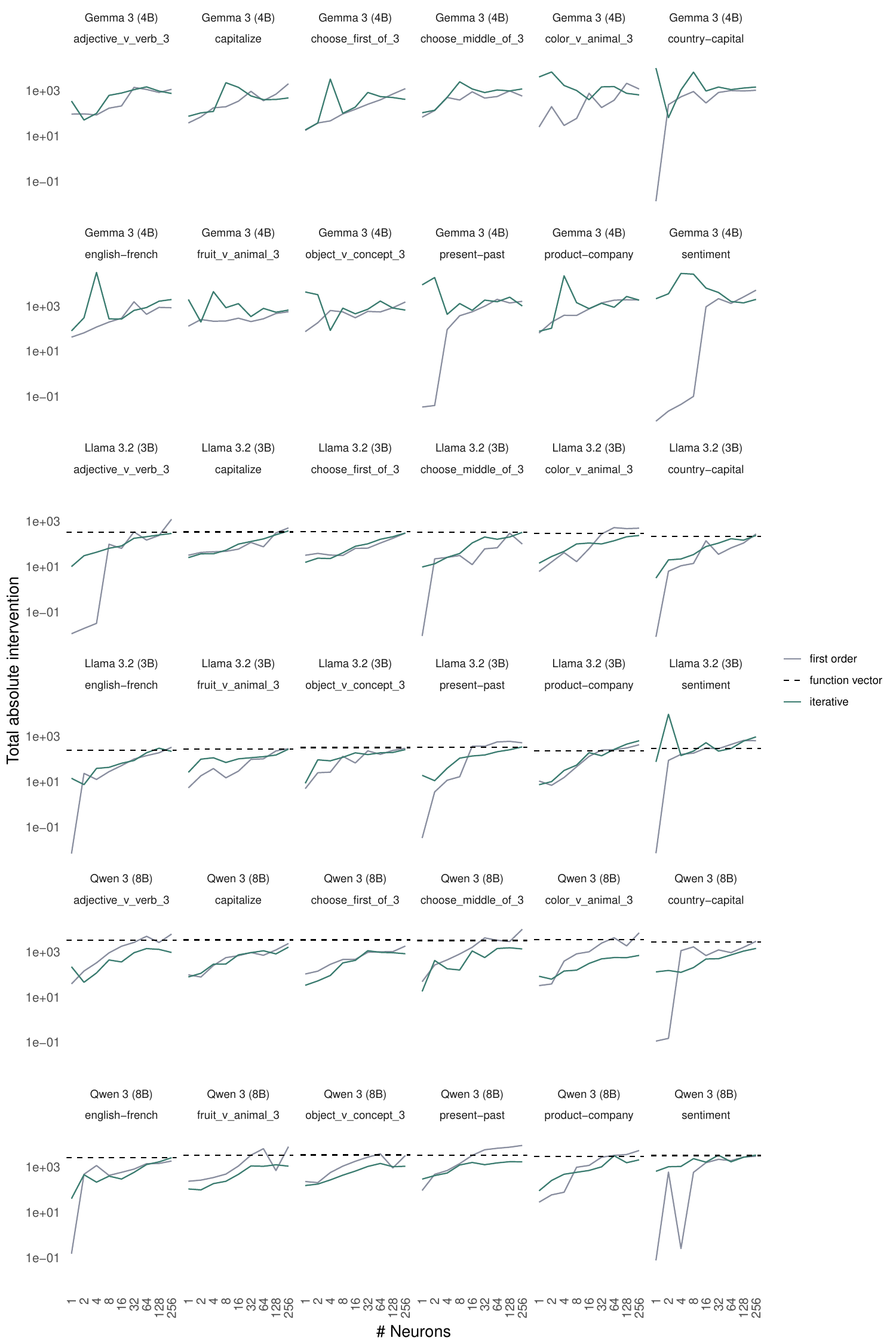}
    \caption{Absolute intervention strength for the first-order, iterative, and function vector interventions, summed over neurons.}
    \label{fig:total}
\end{figure}

\begin{figure}[h]
    \centering
    \includegraphics[width=1\linewidth]{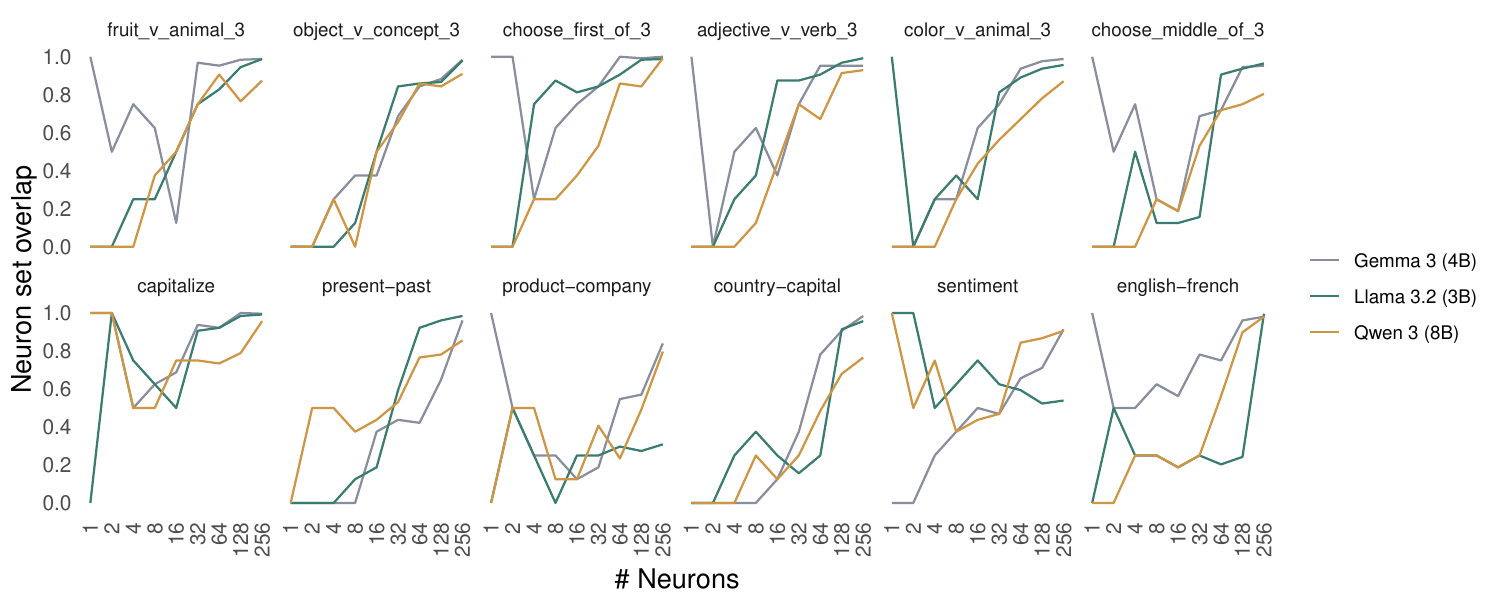}
    \caption{Overlap of neuron sets identified by first-order approximation and the iterative algorithm.}
    \label{fig:overlap}
\end{figure}

\paragraph{Intervention comparison} We briefly analyze the difference in interventions identified by the first-order approximation and the iterative algorithm. Figure \ref{fig:overlap} shows the overlap of the neuron intervention sets of the two intervention methods across tasks and models. While there is overlap, a substantive fraction of the neurons chosen by the iterative procedure are not among the top neurons as identified by the first-order approximation, indicating that the iterative search indeed significantly changes the intervention set. 

Additionally, to verify that differences in task performance between first-order and iterative interventions do not stem from widely different total intervention strengths, rather than differences in the selected neurons, we show in Figure \ref{fig:total} the total intervention strength for the first-order, iterative, and function vector interventions. While there are differences in intervention strength, there is no general trend (such as iterative or first-order interventions always being stronger).

\section{Intervention distribution} \label{sec:pos}

\begin{figure}[h]
    \begin{center}
    \includegraphics[width=.5\textwidth]{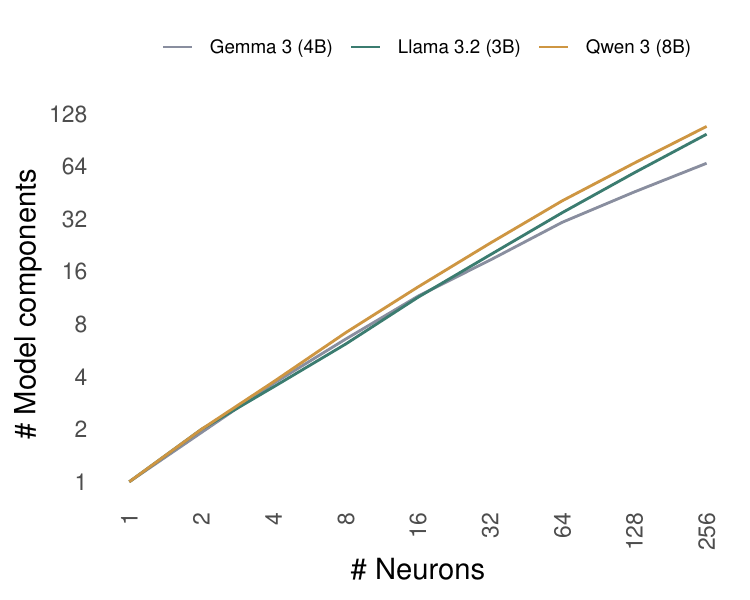}
    \caption{Number of intervened-on model components (attention heads), identified by DSI, per neuron set size, averaged across tasks.}
    \label{fig:components}
    \end{center}
\end{figure}

\begin{figure}[h]
    \centering
    \includegraphics[width=\linewidth]{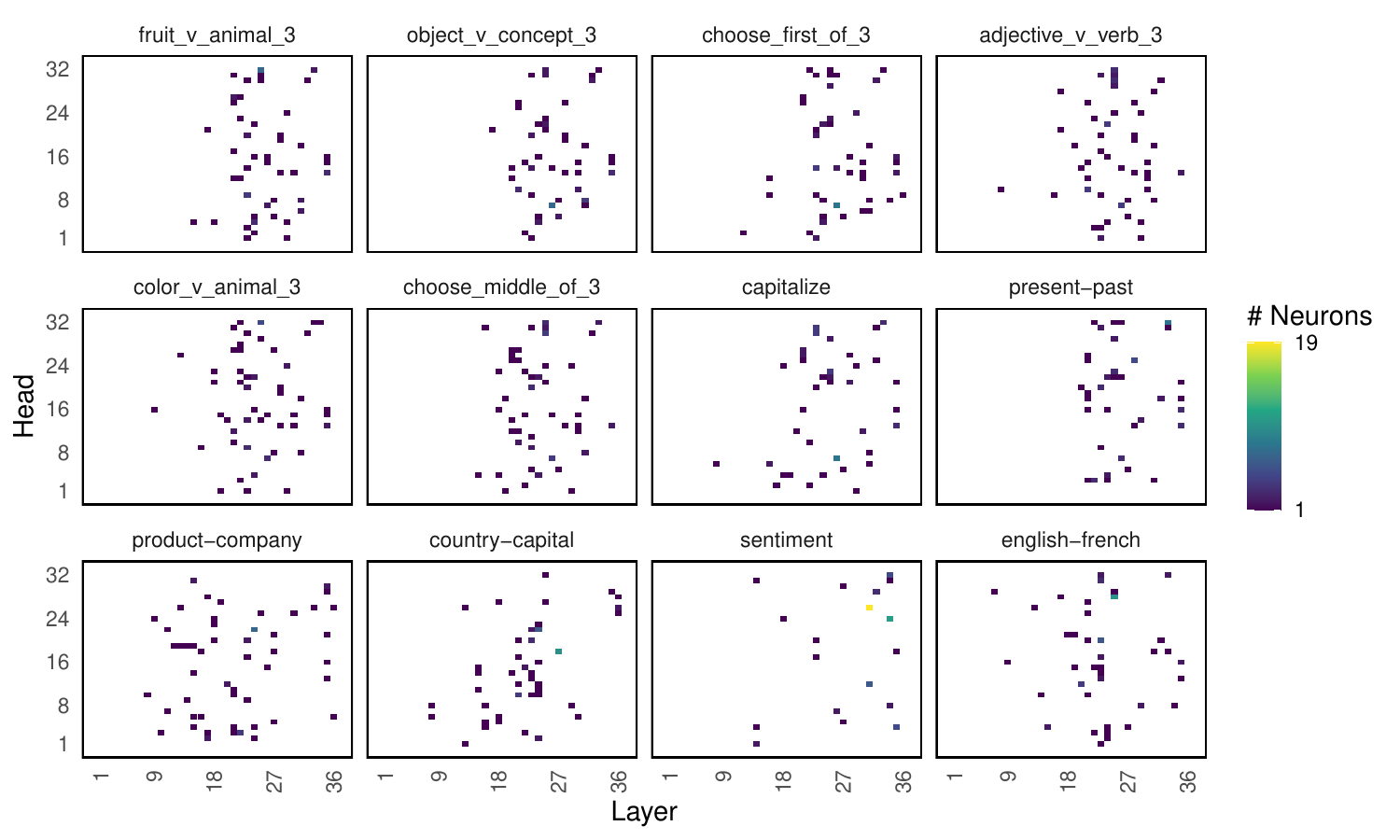}
    \caption{Number of intervened-on neurons, identified by DSI, within layers and heads for Qwen 3 (8B) and a set size of $64$.}
    \label{fig:qwen_neurons}
\end{figure}

\begin{figure}[h]
    \centering
    \includegraphics[width=\linewidth]{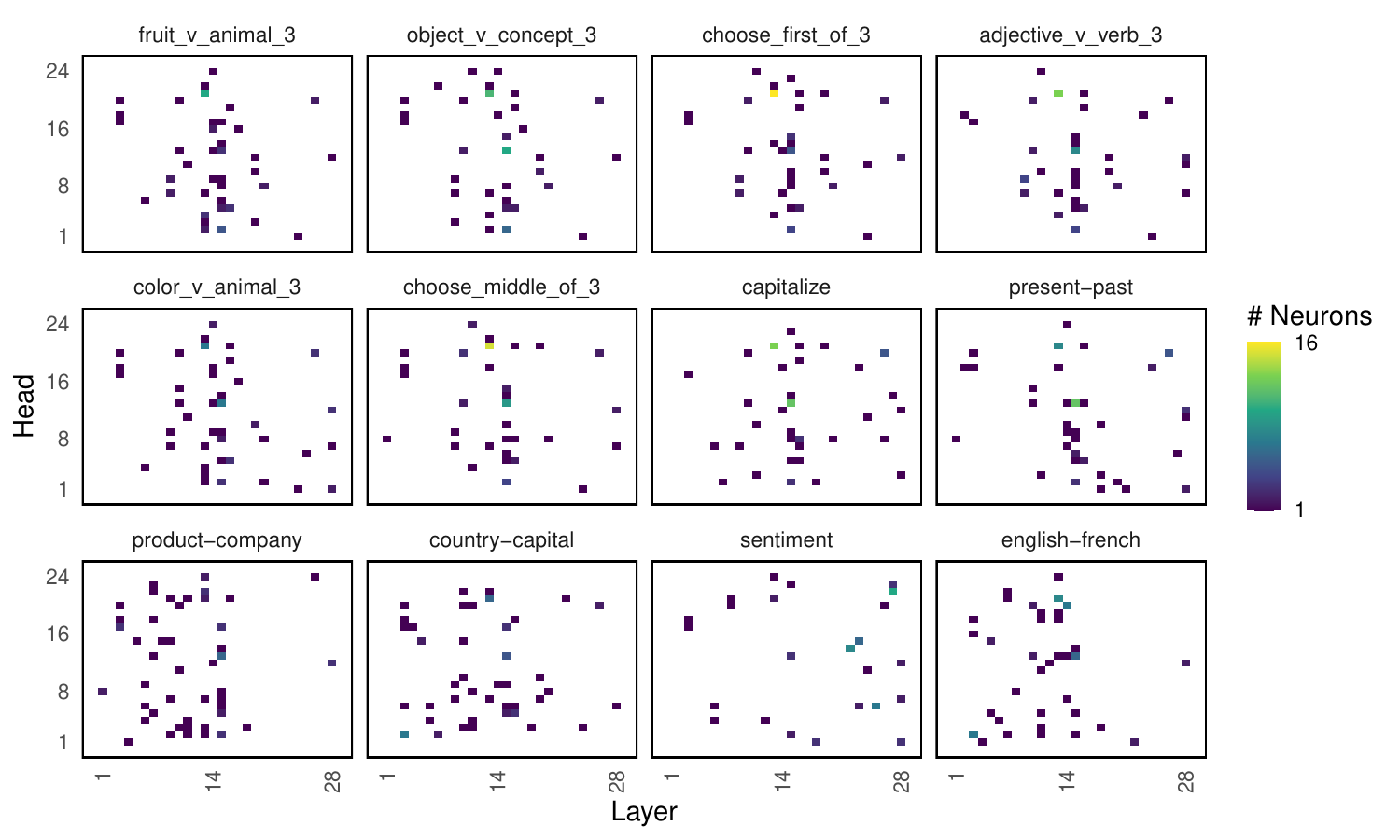}
    \caption{Number of intervened-on neurons, identified by DSI, within layers and heads for Llama 3.2 (3B) and a set size of $64$.}
    \label{fig:llama_neurons}
\end{figure}

\begin{figure}[ht]
    \centering
    \includegraphics[width=\linewidth]{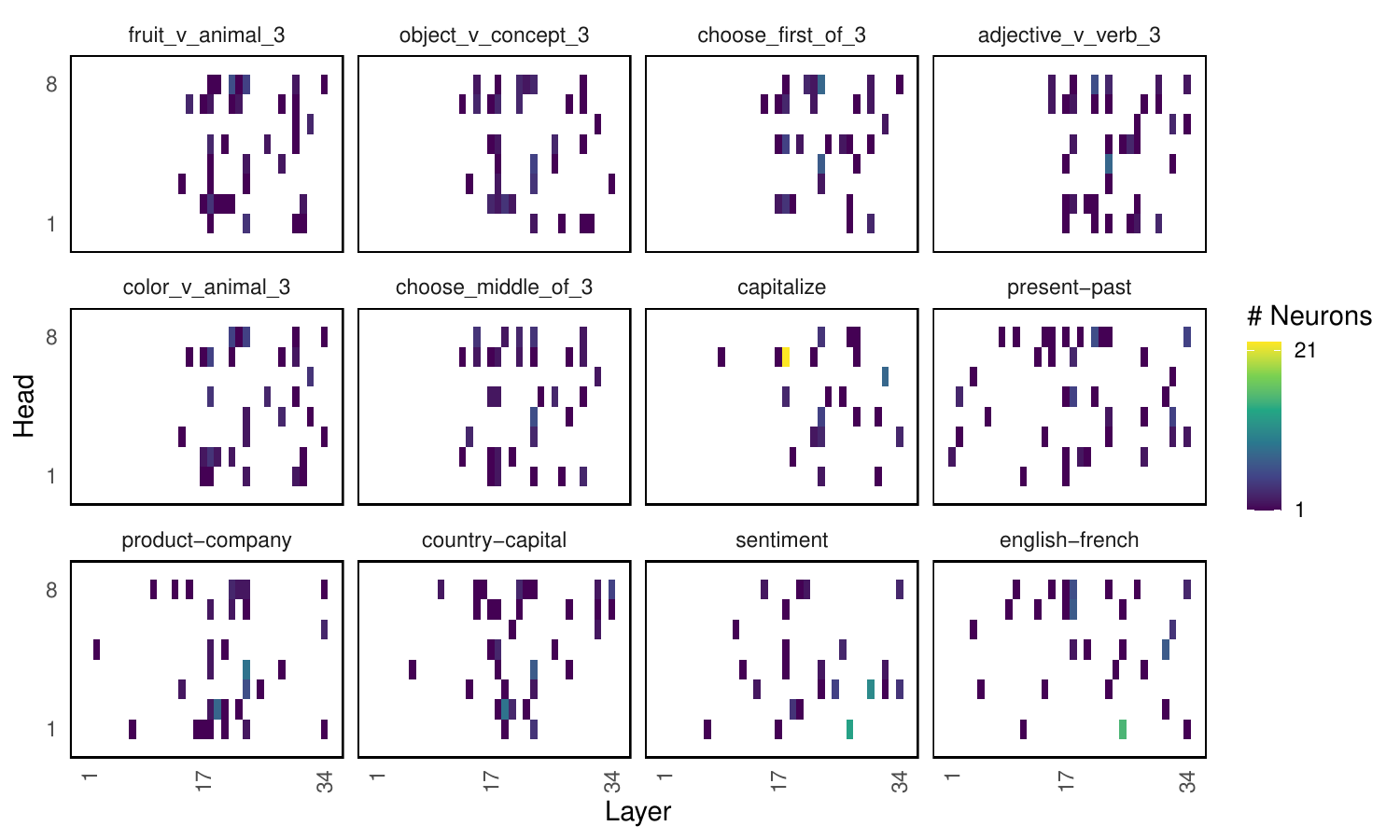}
    \caption{Number of intervened-on neurons, identified by DSI, within layers and heads for Gemma 3 (4B) and a set size of $64$.}
    \label{fig:gemma_neurons}
\end{figure}

In this section, we investigate the distributed nature of DSI interventions, i.e., whether the identified neuron sets distribute widely across layers and attention heads, or whether they are rather tightly localised within the models. Figure \ref{fig:components} shows that DSI interventions, over all tasks, are distributed across many model components (i.e., attention heads), with even very small interventions being spread across multiple heads. Figures \ref{fig:qwen_neurons}, \ref{fig:llama_neurons}, and \ref{fig:gemma_neurons} show the involved layers and heads for DSI interventions of a set size of $64$ for each individual task. The interventions are distributed across layers, with a preference for early-middle to late layers in the Qwen 3 (8B) and Gemma 3 (4B) models.

\newpage \null \newpage

\section{AxBench application details} \label{app:axbench}

\subsection{Adapting DSI to the AxBench setting} \label{app:axbenchalgo}

To compute DSI interventions on AxBench, we again consider the outputs of all attention heads in each layer. There, we compute the average activation difference between the positive and negative examples, and as the objective for the iterative refinement, we compute the cross-entropy with the positive examples.
To save compute, we reduce the number of random restarts to 2 and the number of ZeroFPR iterations to 30.
Since the output consists of multiple tokens now, we average over the output token span to compute activation differences.
For the cross-entropy objective, we use the positive examples as inputs, averaging the cross-entropy of the next token predictions conditional on the prior example tokens (i.e., teacher forcing) over all response-region tokens of an example.
This value is then averaged over the concepts' positive examples.
For steering, we apply the interventions at each token position except for the very first \texttt{<bos>} token.

\subsection{Clustering details} \label{app:clustering}

We perform hierarchical agglomerative clustering with average linkage, using the R function \texttt{hclust}, with 1 - Jaccard index as the distance matrix. To identify clusters from the resulting dendrogram, we use dynamic tree cutting \citep{dynamictree} as implemented in the R package \texttt{dynamicTreeCut}. Since the number of clustered data points is comparatively low, we set the minimum cluster size and the \texttt{deepSplit} parameter to 3, favouring many small clusters, and do not merge unassigned concepts into the nearest cluster. Tables \ref{tab:clusters_2b} and \ref{tab:clusters_9b} contain a full list of identified clusters and their concepts.

\begin{longtable}{rrp{0.75\linewidth}}
\caption{Intervention clusters of Gemma-2-2B for an intervention size of 256; - = not assigned to any cluster.}
\label{tab:clusters_2b} \\
\toprule
Cluster & Score & Concept \\
\midrule
\endfirsthead
\toprule
Cluster & Score & Concept \\
\midrule
\endhead
\bottomrule
\endlastfoot
1 & 1.38 & references to personal pronouns and possessive adjectives \\
 & 1.32 & expressions of enjoyment and fun \\
 & 1.22 & greetings and informal salutations \\
 & 1.20 & statements that express personal beliefs or experiences \\
 & 1.08 & personal pronouns and their usage in relation to self-reference \\
 & 1.04 & sentences that emphasize personal reflections or opinions \\ \midrule
2 & 1.18 & statements, commands, or instructions related to user interactions and comments \\
 & 0.84 & contractions indicating negation \\
 & 0.78 & calls to action for clicking links or buttons \\ \midrule
3 & 1.12 & phrases related to navigating directions and movement \\
 & 0.98 & words related to signals and identifiers in various contexts \\
 & 0.78 & mentions and variations of the word "distance." \\ \midrule
4 & 1.00 & references to folder revisions and their management \\
 & 0.92 & diagrams and schematic representations related to workflows, communication systems, and protocols \\
 & 0.92 & terms related to electrical charge and currents \\
 & 0.84 & quantum-related concepts or terms \\
 & 0.84 & terms and phrases related to servers and server errors \\ \midrule
5 & 1.22 & references to Google and its associated services \\
 & 1.08 & information about events and activities in a community, including announcements, tours, and concerts. \\
 & 0.80 & references to websites and web-related services \\ \midrule
6 & 0.88 & connections and associations in text regarding media influence or misinformation \\
 & 0.84 & language related to controversies and criticism in political contexts \\
 & 0.78 & phrases related to criticism and disapproval towards actions or policies \\ \midrule
7 & 1.22 & references, citations, and numerical data in a text \\
 & 1.12 & detailed references and citations in academic writing \\
 & 1.00 & references to authors or contributors in a scientific context \\ \midrule
8 & 1.12 & terms related to news and its various formats \\
 & 0.98 & elements related to document structure and formatting \\
 & 0.84 & references to sources and reporting in news \\
 & 0.80 & references to figures within a document \\ \midrule
9 & 1.14 & references to legal documents or official communications \\
 & 1.08 & concepts related to legal obligations and consequences \\
 & 0.98 & terms related to trading and financial instruments \\
 & 0.94 & expressions related to legal or financial obligations \\
 & 0.92 & references to legal procedures and rights \\
 & 0.90 & references to legal terms and actions \\
 & 0.90 & terms related to copyrights, licenses, and permissions in software documentation. \\
 & 0.88 & references to temperature measurements and conditions in scientific contexts \\
 & 0.84 & thematic elements related to legal or contractual obligations \\
 & 0.78 & financial terms and concepts related to loans and mortgages \\ \midrule
10 & 1.32 & mentions of significant market factors or trends \\
 & 1.12 & terms related to revenue and financial regulations \\
 & 0.92 & titles and keywords related to economic analysis and market dynamics \\ \midrule
11 & 0.98 & terms related to trade negotiations and governmental decisions \\
 & 0.92 & terms related to governance, citizenship, and roles in societal structures \\
 & 0.84 & titles and roles within academic or professional contexts \\ \midrule
12 & 1.08 & references to brands, products, or specific entities in various contexts \\
 & 1.04 & phrases related to scientific advancements and their implications \\
 & 1.04 & references to media, technology, and medical terminology \\
 & 0.88 & specific patterns related to digital communication and media formats \\ \midrule
13 & 1.22 & phrases related to problem-solving and decision-making challenges \\
 & 1.10 & concepts related to decision-making and management outcomes \\
 & 1.04 & phrases relating to performance expectations and potential improvements in a competitive context \\
 & 0.80 & patterns related to warnings or alerts in a procedural context \\ \midrule
14 & 1.34 & terms related to procedures and techniques \\
 & 1.32 & scientific terms related to research findings and their implications \\
 & 1.14 & scientific concepts and terminology related to uncertainty and measurement \\
 & 0.88 & notes or signals relating to metrics, stats, and data entries \\
 & 0.84 & elements related to scientific data and results \\ \midrule
15 & 1.18 & references to advocacy and support services \\
 & 1.16 & terms related to environmental management and pollution reduction \\
 & 1.04 & terms related to environmental policies and factors \\
 & 0.92 & references to health-related professionals and various stakeholders involved in healthcare discussions \\
 & 0.88 & phrases related to health communication and discussion \\
 & 0.88 & discussions related to educational preparation and career readiness \\
 & 0.80 & key concepts and guidelines related to risk communication in public health research \\ \midrule
16 & 0.96 & references to health-related terms, particularly in the context of potential medical consequences \\
 & 0.92 & references to species or medical terms related to cancer treatments or conditions \\
 & 0.88 & instances of sports-related injuries \\ \midrule
17 & 0.94 & references to time frames and chronological events \\
 & 0.90 & references to political events and figures \\
 & 0.88 & references to significant political events or figures \\
 & 0.80 & historical places or contexts that are significant to specific events or landmarks \\ \midrule
18 & 1.38 & elements related to narrative and storytelling \\
 & 1.20 & elements related to immersive reading experiences and emotional engagement with characters \\
 & 1.04 & references to different worlds, realities, or fantastical settings within narratives \\
 & 0.92 & references to influential artists and creative inspirations in the context of writing and music \\ \midrule
19 & 1.32 & references to family relationships and dynamics \\
 & 1.20 & references to authority figures or personal experiences related to persuasive narratives \\
 & 0.92 & references to children and pediatric contexts \\
 & 0.92 & references to individuals and their experiences or identities \\ \midrule
20 & 1.38 & terms related to personal relationships and social dynamics \\
 & 1.14 & references to relationships and social dynamics \\
 & 1.02 & phrases related to improving relationships and business collaboration \\ \midrule
21 & 1.04 & phrases that describe essential components or elements related to care, sustenance, and values \\
 & 0.98 & phrases related to significant achievements or impactful moments \\
 & 0.92 & words and phrases associated with winning and competition \\
 & 0.88 & phrases related to possession or experiences over time \\ \midrule
22 & 1.14 & concepts related to hidden or concealed elements \\
 & 1.08 & phrases related to awareness and realization \\
 & 1.08 & references to navigation and location tracking \\
 & 1.04 & instances of experiences that are novel or unfamiliar \\
 & 0.92 & references to trips or journeys \\ \midrule
23 & 1.08 & negative language and themes related to distrust and conflict \\
 & 1.02 & concepts related to struggle and self-control, particularly regarding desires and instincts \\
 & 0.94 & key terms and phrases related to turmoil, crises, and significant conflicts in various contexts \\
 & 0.88 & phrases conveying medical conditions and the presence of pain \\ \midrule
24 & 1.60 & phrases that express comparisons or similarities \\
 & 1.04 & names or references to specific individuals or characters \\
 & 1.02 & proper nouns, particularly names and locations \\
 & 0.78 & names of people or entities associated with specific contexts or citations \\ \midrule
25 & 1.38 & phrases that convey necessity or importance \\
 & 1.28 & phrases indicating conditional or preceding circumstances \\
 & 1.22 & conditional phrases indicating necessity or circumstances \\
 & 1.18 & phrases indicating belief, assertion, or opinion about subjects \\
 & 1.14 & conditional phrases starting with "if." \\ \midrule
26 & 1.32 & phrases expressing uncertainty or opinion \\
 & 1.12 & phrases that express skepticism or critical viewpoints \\
 & 0.98 & statements about opinions, assertions, and disclaimers regarding information and its accuracy \\ \midrule
27 & 1.12 & instances of the word "next." \\
 & 1.08 & phrases that indicate the beginning of actions or events \\
 & 0.96 & phrases related to repetitive actions or follow-up events \\
 & 0.90 & actions related to movement or directives, particularly in tense contexts \\ \midrule
28 & 1.22 & terms related to maintaining or preserving a state or condition \\
 & 1.16 & qualifiers and intensifiers that modify the degree of an adjective or adverb \\
 & 1.08 & statements of existence or presence \\
 & 1.02 & various types of items or features, often describing characteristics or attributes \\
 & 1.00 & the presence of the word "public." \\
 & 0.94 & terms related to differentiation, separation, and distinction \\
 & 0.92 & verbs that indicate ease or improvement in processes or experiences \\
 & 0.88 & terms related to adaptability and adjusting to changing conditions \\
 & 0.84 & words and phrases related to service offerings and product availability \\ \midrule
29 & 1.44 & instances of comparisons and contrasts in statements \\
 & 1.16 & conjunctions and linking words that facilitate the connection between ideas \\
 & 1.12 & words and phrases indicating comparison or qualification \\
 & 1.02 & the word "such" in various contexts, often indicating conditions or examples \\
 & 1.02 & phrases indicating comparisons or relationships between two entities or elements \\
 & 0.92 & the word "much" in various forms and contexts \\ \midrule
-- & 1.44 & questions beginning with "how." \\
 & 1.28 & phrases related to providing information or guidance about procedures \\
 & 1.22 & references to candy and confectionery products \\
 & 1.14 & references to menus and related menu items within content \\
 & 1.10 & technical or operational terms related to project management and documentation \\
 & 0.98 & references to farms and agricultural concepts \\
 & 0.98 & references to standardized testing \\
 & 0.96 & punctuation marks, particularly semicolons and parentheses \\
 & 0.94 & contextual details concerning costs and pricing \\
 & 0.92 & the presence of single quotation marks \\
 & 0.92 & descriptive elements related to architecture and physical spaces \\
 & 0.92 & numerical citations or references within the text \\
 & 0.84 & specific formatting or markers in the text structure \\
 & 0.84 & elements related to procedural guidelines and recommendations \\
 & 0.84 & factors related to communication system performance and scientific research needs \\
 & 0.80 & references to specific teams and player statistics in sports \\
\end{longtable}

\begin{longtable}{rrp{0.75\linewidth}}
\caption{Intervention clusters of Gemma-2-9B for an intervention size of 256; - = not assigned to any cluster.}
\label{tab:clusters_9b} \\
\toprule
Cluster & Score & Concept \\
\midrule
\endfirsthead
\toprule
Cluster & Score & Concept \\
\midrule
\endhead
\bottomrule
\endlastfoot
1 & 1.32 & elements related to validation and classification criteria in a structured data context \\
 & 1.22 & terms related to accumulation and buildup of substances \\
 & 1.20 & sentence-ending punctuation and transitional phrases \\
 & 1.12 & references to security-related terms \\
 & 1.12 & references to real estate and property ownership \\
 & 1.08 & past tense verbs indicating actions or events \\
 & 1.04 & references to educational background and institutions \\
 & 0.96 & references to geographical locations, specifically cities \\
 & 0.92 & references to concealed or enclosed spaces and protective casings \\
 & 0.92 & terms related to winning or achieving victories in competitions or games \\
 & 0.84 & references to relationships and joint decisions \\
 & 0.84 & components related to educational sessions and presentations \\
 & 0.80 & references to big data technologies and hadron physics \\
 & 0.78 & technical specifications and requirements related to software and hardware \\ \midrule
2 & 1.64 & words that indicate personal pronouns or first-person references \\
 & 1.38 & conversational phrases that indicate problem-solving or assistance \\
 & 1.14 & instances of direct speech or quotation \\
 & 1.02 & phrases indicating instructions or tutorials involving the word "to." \\
 & 0.98 & the presence of personal pronouns and certain numerical values in the text \\ \midrule
3 & 1.26 & references to divine beings and celestial objects \\
 & 1.10 & references to nighttime or nocturnal themes \\
 & 0.84 & words and phrases related to elegance and elaborate style \\ \midrule
4 & 0.98 & expressions of vulnerability and honesty about personal struggles \\
 & 0.94 & expressions of personal experience and struggles related to judgment and stigma \\
 & 0.80 & themes related to personal loss and health struggles \\
 & 0.78 & references to difficult or distressing situations and the associated emotional responses \\ \midrule
5 & 1.32 & elements of personal experience and emotional state \\
 & 1.18 & pronouns and references to the self or people in a narrative context \\
 & 1.08 & phrases related to past events and personal connections \\ \midrule
6 & 1.48 & terms related to personal interests and emotional experiences \\
 & 0.94 & positive sentiments and descriptions of enjoyable experiences \\
 & 0.84 & emotional expressions and responses in relation to movies \\
 & 0.84 & phrases that contain references to the concept of art or artistic content \\
 & 0.84 & various expressions of emotional or sensory perceptions \\ \midrule
7 & 1.32 & references to musical instruments or compositions \\
 & 1.22 & phrases and concepts related to gastronomy and culinary experiences \\
 & 1.16 & references to food items and catering options \\
 & 1.12 & phrases related to visiting or engaging with specific activities or locations \\
 & 1.08 & references to social gatherings and events \\
 & 1.04 & references to color and backgrounds in design contexts \\
 & 0.92 & terms related to online gambling and casinos \\
 & 0.84 & references to music and songs \\
 & 0.84 & references to dressing or attire \\ \midrule
8 & 1.22 & technical terms and phrases related to scientific measurements and research methodologies \\
 & 1.16 & phrases related to measurement processes and specific scientific terms \\
 & 0.84 & specific terms and phrases related to scientific protocols and methodologies \\
 & 0.78 & terms related to academic research, databases, and technical fields \\ \midrule
9 & 1.02 & specific terms related to efficiency and regulation within systems \\
 & 0.94 & key structural elements and functions within technical or scientific contexts \\
 & 0.88 & specific terms related to technical aspects and functionalities in various contexts \\ \midrule
10 & 1.30 & keywords related to scientific testing and chemical analysis \\
 & 0.94 & technical vocabulary related to semiconductor components and processes \\
 & 0.92 & technical terminology and data related to particle physics experiments \\
 & 0.88 & terms and concepts related to biology and medicine \\
 & 0.78 & terms related to biochemical compounds and their effects \\
 & 0.78 & terms related to neuroscience and brain function \\ \midrule
11 & 1.18 & terms related to energy transfer and conversion processes \\
 & 0.88 & references to "Digital Twins" and related concepts \\
 & 0.84 & terms and concepts associated with cloud computing and virtualization services \\ \midrule
12 & 1.38 & words or phrases related to achievement and effectiveness \\
 & 1.28 & intensifiers that emphasize magnitude or degree \\
 & 1.26 & words associated with importance or significance \\
 & 0.98 & expressions related to admiration and respect for individuals \\ \midrule
13 & 1.10 & phrases that express awareness or knowledge about a situation \\
 & 1.08 & references to the user or their actions \\
 & 1.04 & sentences reflecting understanding and agreement in discussions \\ \midrule
14 & 1.32 & phrases indicating future events or actions \\
 & 0.96 & instances of the word "provided" and its variations \\
 & 0.94 & phrases indicating obligation or necessity \\ \midrule
15 & 1.22 & elements that indicate comparisons or contrasts between subjects \\
 & 1.22 & phrases indicating causation and the effects of specific actions or conditions \\
 & 1.22 & instances of the word "including." \\
 & 1.14 & different forms of combinations or integrations between multiple elements or factors \\
 & 1.08 & phrases that indicate additional information or context being included \\
 & 1.02 & phrases indicating relationships or associations within a context \\
 & 0.96 & aspects of clarification and detail in communication \\
 & 0.94 & terms indicating the quality or effectiveness of performance and health outcomes \\
 & 0.84 & information related to categories or classifications \\ \midrule
16 & 1.22 & instances of the word "show" and its variations \\
 & 1.16 & phrases indicating frequency and patterns of behavior \\
 & 1.08 & prepositions and phrases of location and context \\
 & 1.08 & the term "wide" in various contexts \\
 & 0.92 & references to physical contact or proximity between objects or people \\
 & 0.92 & repetitive sequences or patterns in data \\
 & 0.92 & instances of the word "face" and its variations, indicating confrontations or challenges faced by subjects in the text \\
 & 0.84 & phrases related to totality or absence \\
 & 0.80 & instances of the term "success" \\
 & 0.80 & occurrences of the word "on" \\ \midrule
17 & 1.42 & phrases that emphasize inclusion and represent a sense of belonging \\
 & 1.18 & expressions of commitment to collaboration and community responsibility \\
 & 1.12 & concepts and discussions related to creativity and innovation \\
 & 1.08 & terms related to outreach and audience engagement \\
 & 0.92 & terms related to social organizations and community involvement \\ \midrule
18 & 1.12 & references to churches and religious institutions \\
 & 0.96 & references to significant events or actions related to time and people \\
 & 0.94 & references to different religious affiliations and conditions related to mental health \\
 & 0.94 & terms related to voting and election processes \\
 & 0.88 & references to power dynamics and political figures \\
 & 0.84 & terms related to mental health issues and emotional distress \\ \midrule
19 & 1.18 & references to templates and user-generated content creation processes \\
 & 1.00 & links and references to accounts and privacy-related content \\
 & 0.88 & terms related to Creative Commons licensing \\ \midrule
20 & 1.28 & references to businesses and industries \\
 & 1.08 & references to specific individuals or entities involved in reported events \\
 & 0.98 & references to people or groups in a social context \\
 & 0.96 & mentions of institutional or organizational affiliations \\
 & 0.92 & references to government departments or agencies \\
 & 0.90 & references to specific geographic locations and areas in the context of economic discussion \\
 & 0.84 & mentions of government officials and institutional affiliations \\ \midrule
21 & 1.12 & references to reports, surveys, or studies that include data and conclusions \\
 & 1.04 & references to academic journals and publications \\
 & 0.98 & references to data sets and comparative analysis in research contexts \\
 & 0.94 & references to studies or data related to medical conditions and their treatments \\
 & 0.80 & data and findings related to research outcomes and analyses \\ \midrule
22 & 1.24 & terms and concepts related to specific fields, particularly in law, health, and technology \\
 & 1.04 & legal terminology and references to specific cases or patients \\
 & 0.98 & keywords related to legal and political concepts \\
 & 0.84 & phrases related to political and legal context \\
 & 0.84 & references to legal proceedings and court orders \\ \midrule
23 & 1.32 & terms and phrases related to transaction processes and review mechanisms \\
 & 1.18 & references to systems and their components \\
 & 1.12 & terms related to eligibility and selection processes \\ \midrule
24 & 1.08 & phrases indicating a need for improvement or development in technology \\
 & 0.92 & references to education and training backgrounds \\
 & 0.88 & terminology related to restrictions and health measures \\
 & 0.88 & references to changes in rules and policies affecting guidelines or regulations \\ \midrule
25 & 1.26 & the word "either" in various contexts \\
 & 1.18 & questions and statements challenging beliefs or assumptions \\
 & 0.98 & various types of conditional and contrasting phrases \\ \midrule
26 & 1.54 & specific punctuation or words that indicate enthusiasm or emphasis \\
 & 1.50 & imperative phrases and calls to action \\
 & 1.38 & informal second-person pronouns and abbreviations commonly used in casual communication \\
 & 1.32 & playful and onomatopoeic sounds or expressions \\
 & 1.22 & expressions of surprise or disbelief \\
 & 0.98 & conversational expressions or discourse markers \\ \midrule
-- & 1.38 & inquiries and questions about objects, actions, or concepts in various contexts \\
 & 1.32 & references to internet links or URLs \\
 & 1.28 & elements related to data visualizations and plots \\
 & 1.12 & terms related to food and energy prices \\
 & 1.08 & terms related to defects and corrective actions in reports \\
 & 1.08 & occurrences of significant transitional phrases and markers \\
 & 1.02 & phrases related to concerns and the consequences of inaction \\
 & 0.98 & references to specific items or products related to storage solutions and organization \\
 & 0.98 & phrases related to formal requests and demands \\
 & 0.96 & recommendations for actions, particularly regarding media and activities \\
 & 0.90 & terms related to research methodologies and educational frameworks \\
 & 0.80 & proper nouns and specific names related to characters, objects, or places \\
 & 0.80 & formal language related to methodology and research findings \\
 & 0.78 & elements within structured lists or updates \\
 & 0.78 & phrases related to service and customer support \\
\end{longtable}

\end{document}